\documentclass[lettersize,journal]{IEEEtran}
\usepackage{amsmath,amsfonts}
\usepackage{algorithm}
\usepackage{algpseudocode} 
\usepackage{array}
\usepackage[caption=false,font=normalsize,labelfont=sf,textfont=sf]{subfig}
\usepackage{textcomp}
\usepackage{stfloats}
\usepackage{url}
\usepackage{verbatim}
\usepackage{cite}
\usepackage{graphicx}
\usepackage{amsmath,amssymb}
\usepackage{booktabs}
\usepackage{multirow}
\usepackage{colortbl}
\usepackage{pifont}  

\usepackage{float}
\usepackage{adjustbox}  
\usepackage[table,xcdraw]{xcolor}
\usepackage{ulem}
\usepackage[pagebackref,breaklinks,colorlinks]{hyperref}
\usepackage{cleveref}
\crefname{figure}{Fig.}{Figs.}
\Crefname{figure}{Fig.}{Figs.}
\usepackage{ragged2e}
\usepackage{afterpage} 
\usepackage{placeins}
\graphicspath{{figures/}}
\newcommand{\etal}{\textit{et al.}}
\definecolor{darkgreen}{RGB}{0,176,80}  
\definecolor{darkred}{RGB}{220,0,0}   

\usepackage{soul, color, xcolor}
\definecolor{myyellow}{RGB}{255,255,0}
\sethlcolor{myyellow}  
\newif\ifhighlight
\highlightfalse     
\newcommand{\hlA}[1]{\ifhighlight\hl{#1}\else#1\fi}

\newif\ifhighlightB   
\highlightBfalse     
\newcommand{\hlB}[1]{\ifhighlightB\hl{#1}\else#1\fi}

\soulregister{\textbf}7
\soulregister{\textit}7
\soulregister{\cite}7 
\soulregister{\cref}7 
\soulregister{\Cref}7  
\soulregister{\underline}7 %

\begin{document}

\title{Why and How: Knowledge-Guided Learning for Cross-Spectral \\ Image Patch Matching}

\author{Chuang Yu, Yunpeng Liu*, Jinmiao Zhao, Ao Chen, Xiujun Shu, Bo Wang,\\ Zelin Shi, and Xiangyu Yue*
\thanks{This research was supported by the LiaoNing Revitalization Program under Grant no. XLYC2201001. \textit{(Corresponding author: Yunpeng Liu, Xiangyu Yue.)}}
\thanks{Chuang Yu and Jinmiao Zhao are with the Key Laboratory of Opto-Electronic Information Processing and the Shenyang Institute of Automation, Chinese Academy of Sciences, Shenyang 110016, China, and also with the School of Computer Science and Technology, University of Chinese Academy of Sciences, Beijing 100049, China (e-mail:  yuchuang@sia.cn; zhaojinmiao@sia.cn).}
\thanks{Yunpeng Liu and Zelin Shi are with the Shenyang Institute of Automation, Chinese Academy of Sciences, Shenyang 110016, China (email: ypliu@sia.cn; zlshi@sia.cn)}
\thanks{Ao Chen is with the State Key Laboratory of
Mechanical System and Vibration, Shanghai Jiao Tong University, Shanghai 200240, China (e-mail:llykevin@sjtu.edu.cn).}
\thanks{Xiujun Shu and Bo Wang are with the Tencent, Shenzhen 518054, China (e-mail:xiujunshu@tencent.com; pollowang@tencent.com).}
\thanks{Xiangyu Yue is with the MMLab of Chinese University of Hong Kong, Hong Kong 999077, China. (email: xyyue@cuhk.edu.hk)}}

\markboth{IEEE Transactions on Image Processing}%
{Shell \MakeLowercase{\textit{et al.}}: A Sample Article Using IEEEtran.cls for IEEE Journals}


\maketitle

\begin{abstract}
Recently, cross-spectral image patch matching based on feature relation learning has attracted extensive attention. However, existing methods focus on mining richer feature relations by building complex relation extraction structures. Meanwhile, \hlB{performance bottlenecks have gradually emerged.} To address this, we make the first attempt to explore a stable and efficient bridge between descriptor learning and metric learning, and construct a \underline{K}nowledge-\underline{G}uided \underline{L}earning \underline{Net}work (KGL-Net), which achieves \hlA{significant} performance improvements while abandoning complex network structures. Specifically, we find that there is feature extraction consistency between metric learning based on feature difference learning and descriptor learning based on Euclidean distance. This provides the foundation for bridge building. To ensure the stability and efficiency of the constructed bridge, on the one hand, we conduct an in-depth exploration of 20 combined network architectures. On the other hand, a feature-guided loss is constructed to achieve mutual guidance of features. In addition, unlike existing methods, we consider that the feature mapping ability of the metric branch should receive more attention. Therefore, a hard negative sample mining for metric learning (HNSM-M) strategy is constructed. To the best of our knowledge, this is the first time that hard negative sample mining for metric networks has been implemented and brings significant performance gains. Extensive experimental results show that our KGL-Net achieves SOTA performance in \hlB{multiple} cross-spectral image patch matching datasets. Our code is available at \href{https://github.com/YuChuang1205/KGL-Net}{https://github.com/YuChuang1205/KGL-Net}.
\end{abstract} 


\vspace{5pt}

\section{Introduction}
\IEEEPARstart{I}{mage} patch matching aims to measure the similarity between image patches and is widely used in image registration~\cite{yu2023feature,quan2022deep,quan2022self}, image retrieval~\cite{lin2023zero,balntas2017hpatches,shi2022beyond}, and re-identification~\cite{wang2018person,chen2022structure,yang2019patch}. Considering that a single-spectral image source often has limitations, cross-spectral image matching can establish the correspondence between different spectral images, laying the foundation for fully utilizing the complementary information~\cite{aguilera2016learning,ofir2018deep}. Compared with single-spectral image patch matching, cross-spectral image patch matching not only needs to \hlA{face} illumination changes and geometric changes, but also has to overcome the pixel-level nonlinear differences between cross-spectral image patches~\cite{yu2023feature,quan2019afd,quan2021element,yu2022multibranch,yu2023efficient}. Therefore, it is challenging to construct a high-performance cross-spectral image patch matching method.

\begin{figure}[t]
  \captionsetup{skip=2pt}
  \centering
   \includegraphics[width=\columnwidth]{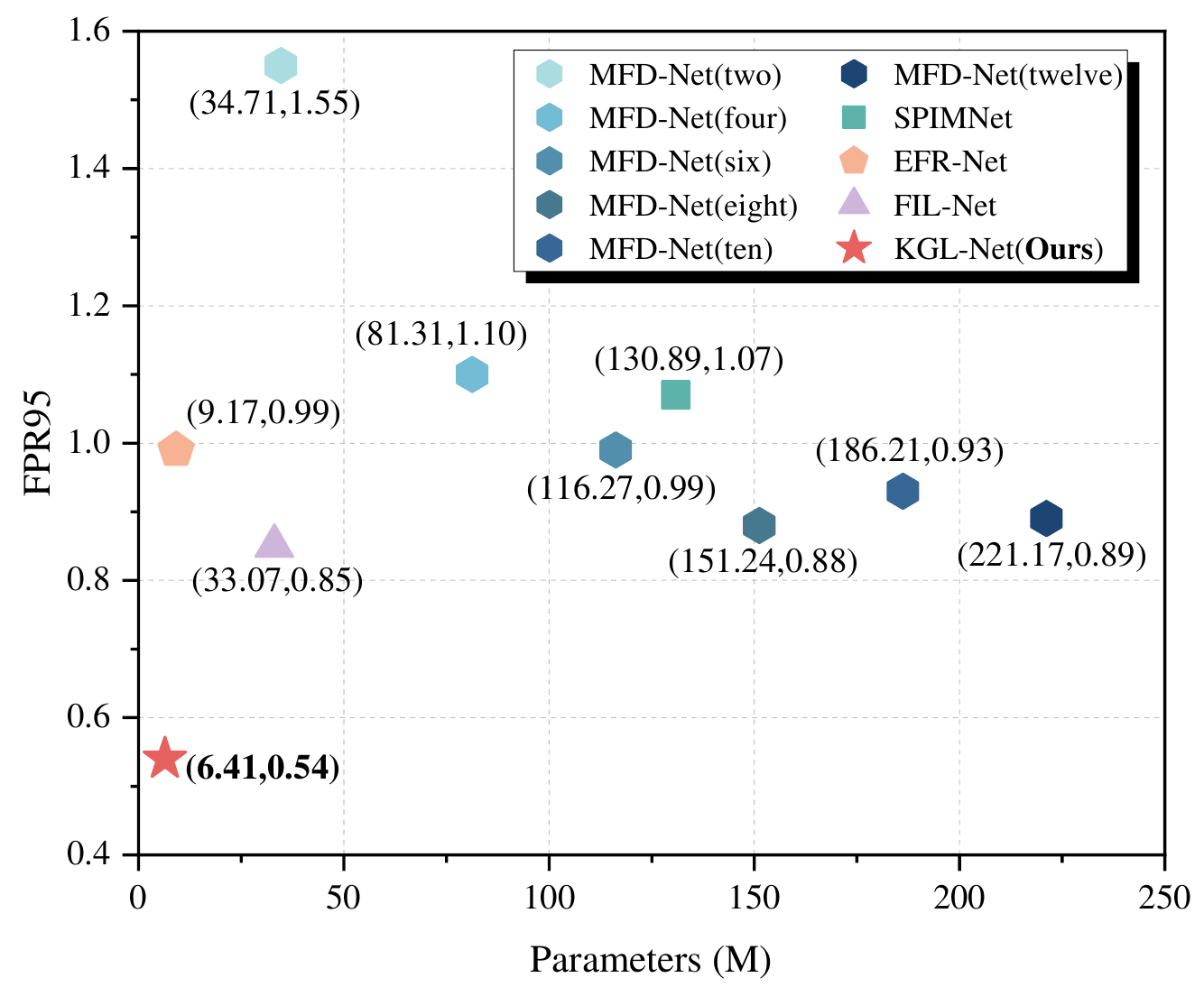}
   \vspace{-15pt}
   \caption{Comparison of different cross-spectral image patch matching methods on the VIS-NIR patch dataset. Our KGL-Net achieves \hlA{excellent} performance. Compared with the latest FIL-Net~\cite{yu2023feature}, KGL-Net reduces the FPR95 by \textbf{36.5\%} (from 0.85 to 0.54). Compared with the lightweight EFR-Net~\cite{yu2023efficient}, KGL-Net reduces the parameters by \textbf{30.1\%} (from 9.17 to 6.41).}
   \label{fig:fig01}
   \vspace{-5pt}
\end{figure}

Early image patch matching methods are based on non-deep learning methods, which mainly study grayscale information~\cite{barnea1972class,azriel1977coarse,vanderbrug1977two,viola1997alignment} or hand-crafted descriptors~\cite{lowe2004distinctive,dalal2005histograms,bay2008speeded,rublee2011orb,firmenichy2011multispectral,pinggera122012cross,morris2007statistics,aguilera2012multispectral,alcantarilla2012kaze,mouats2014multispectral,aguilera2015lghd}. The former studies image pixels, and the latter studies image gradients. However, this type of method is seriously affected by hyperparameters and lacks robustness for complex transformation scenarios. Compared with model-driven non-deep learning-based methods, deep learning-based methods can automatically learn the nonlinear mapping between input images and true labels~\cite{lecun2015deep,zhao2024hierarchical,chen2025discrete,hu2025cost,zhao2025multi}, which has greater adaptability. Deep learning-based methods can be divided into descriptor learning methods~\cite{simo2015discriminative,balntas2016learning,balntas2016pn,kumar2016learning,aguilera2017cross,zhang2017learning,mishchuk2017working,wang2017train,zheng2019hardness,wang2019better,tian2017l2,tian2019sosnet,tian2020hynet} and metric learning methods~\cite{yu2023feature,aguilera2016learning,quan2019afd,quan2021element,yu2022multibranch,yu2023efficient,han2015matchnet,zagoruyko2015learning,quan2021multi,quan2018cross,en2018ts,ko2023spectral}. The descriptor learning method uses a deep learning network to extract high-level features of the input image patch and measures the similarity between feature descriptors by feature distance. Different from the descriptor learning method, the metric learning method is an end-to-end network that extracts and measures image patch features in sequence by using a neural network. Owing to the structural characteristics of these two methods, research on descriptor learning methods focuses on optimizing loss function~\cite{simo2015discriminative,balntas2016learning,balntas2016pn,kumar2016learning,aguilera2017cross,zhang2017learning,wang2019better,tian2017l2,tian2019sosnet,tian2020hynet} and hard sample mining~\cite{mishchuk2017working,wang2017train,zheng2019hardness,tian2017l2}, whereas research on metric learning methods focuses on the design of network structures to mine more sufficient and comprehensive feature relations between image patches~\cite{yu2023feature}. The former has the characteristics of lightweight network but low accuracy. The latter has the characteristics of high accuracy but bloated network. \hlB{Considering that matching accuracy is treated as the priority metric}, existing cross-spectral image patch matching methods focus on metric learning~\cite{yu2023feature,quan2019afd,quan2021element,yu2022multibranch,yu2023efficient}. However, as shown in \cref{fig:fig01}, although researchers have made great efforts in recent years, the research idea of focusing solely on building more complex feature relation extraction network structures to mine rich feature relations makes further performance improvements increasingly challenging. How to break through this bottleneck has become an important challenge in current research.

\hlB{To address this, we consider that it is necessary to break away from the inertial thinking of building more complex feature relation extraction network structures and explore a new paradigm.} Combining the advantages of descriptor learning~\cite{tian2020hynet}, we make the first attempt to build a stable and efficient bridge between descriptor learning and metric learning. During the exploration, we find that there is feature extraction consistency between metric learning based on feature difference learning and descriptor learning based on Euclidean distance. They all focus on extracting the differential features of the image patch itself relative to other image patches. Based on this and to ensure the stability and efficiency of the bridge built, on the one hand, we \hlB{systematically explore 20 combined network architectures} of descriptor learning and metric learning. On the other hand, we construct a feature-guided loss to achieve effective mutual guidance of features. \hlB{In addition, unlike existing methods that focus on the feature extractor design, we argue that the feature mapping ability of the metric branch should also receive more attention.} Considering that the metric branch is essentially a learnable discriminator, its ability is directly related to the input training data. \hlB{Therefore, we make the first attempt to design an effective hard negative sample mining strategy for metric learning.} On the one hand, we break the existing metric learning methods that directly use uniform and fixed positive and negative sample pairs, and use only positive sample pairs as inputs. This provides the basic conditions for hard negative sample mining. On the other hand, considering that the metric network uses the metric branch to combine the two-branch features~\cite{han2015matchnet,zagoruyko2015learning}, it is unrealistic to directly perform hard negative sample mining. \hlB{Therefore, we attempt to leverage descriptor learning to provide effective knowledge guidance. Based on this, we construct a hard negative sample mining for metric learning (HNSM-M) strategy, which uses the feature distance matrix generated by descriptor learning to guide  effective hard negative sample mining in metric learning. The contributions of this paper are summarized as follows:}

\begin{itemize}
    \item To solve the performance bottleneck in the current research, we make the first attempt to explore a stable and efficient bridge between descriptor learning and metric learning and construct a KGL-Net, which achieves \hlA{significant} performance improvements while abandoning complex network structures.
    \item We find that there is feature extraction consistency between metric learning based on feature difference learning and descriptor learning based on Euclidean distance. Based on this and to ensure the stability and efficiency of the constructed bridge, \hlB{we not only conduct an in-depth exploration of 20 combined network architectures, but also construct a feature-guided loss to achieve mutual guidance of feature extraction.}
    \item We focus on improving the feature mapping capability of the metric branch and construct a HNSM-M strategy. To the best of our knowledge, this is the first time that hard negative sample mining for metric networks is implemented and brings significant performance gains.
\end{itemize}

The rest of the manuscript is organized as follows: \Cref{sec:related_work} briefly reviews the \hlB{related work}. \Cref{sec:methods} provides a detailed description of KGL-Net and its components. \Cref{sec:experiments} presents the experimental setup, results, and analysis. \Cref{sec:conlusion} concludes the paper.

\vspace{10pt}
\section{Related Work}
\label{sec:related_work}

 The research on image patch matching methods can be divided into non-deep learning-based methods and deep learning-based methods. Deep learning-based methods can be further divided into descriptor learning methods and metric learning methods. Therefore, this section will briefly review non-deep learning-based methods, descriptor learning methods, and metric learning methods.

\subsection{Non-deep learning-based methods}

Early image patch matching methods mainly focused on single-spectral images, relying on grayscale features~\cite{barnea1972class,azriel1977coarse,vanderbrug1977two,viola1997alignment} and hand-crafted feature descriptors~\cite{lowe2004distinctive,dalal2005histograms, bay2008speeded, rublee2011orb, alcantarilla2012kaze}. The former methods measure the similarity between image patches by directly computing metrics such as mean absolute difference (MAD)~\cite{barnea1972class}, normalized cross-correlation (NCC)~\cite{azriel1977coarse,vanderbrug1977two}, and mutual information (MI)~\cite{viola1997alignment}. In contrast, the latter methods extract local gradient histogram features and perform patch matching based on feature space distances. Representative algorithms include scale-invariant feature transform (SIFT)~\cite{lowe2004distinctive}, histogram of oriented gradients (HOG)~\cite{dalal2005histograms}, speeded-up robust features (SURF)~\cite{bay2008speeded}, and oriented FAST and rotated BRIEF (ORB)~\cite{rublee2011orb}. Compared with single-spectral images, cross-spectral images have significant pixel-level nonlinear differences. Traditional methods have limited adaptability in this context, so related research started relatively late. To address this challenge, early research extended feature modeling methods from single-spectral matching to cross-spectral scenarios by integrating cross-spectral characteristics, focusing on robust representations of gradient orientations and edge structures across spectral domains. In terms of gradient feature extraction,~\cite{firmenichy2011multispectral} and~\cite{pinggera122012cross} respectively employed SIFT and HOG descriptors to enhance the robustness of cross-spectral matching against gradient variations. In terms of edge feature modeling, considering the consistency of object boundaries in cross-spectral images~\cite{morris2007statistics},~\cite{aguilera2012multispectral} proposed the local edge histogram descriptor (EHD). Subsequently,~\cite{mouats2014multispectral} and~\cite{aguilera2015lghd} further improved the EHD method by introducing multi-scale Log-Gabor filter responses to enhance the feature expression capability. However, due to the drastic changes in cross-spectral imaging, non-deep learning-based methods still have obvious shortcomings in adaptability and robustness, and further breakthroughs are urgently needed.

\begin{figure*}[htbp]
    \captionsetup{skip=3pt}
    \centering
    \includegraphics[width=\textwidth]{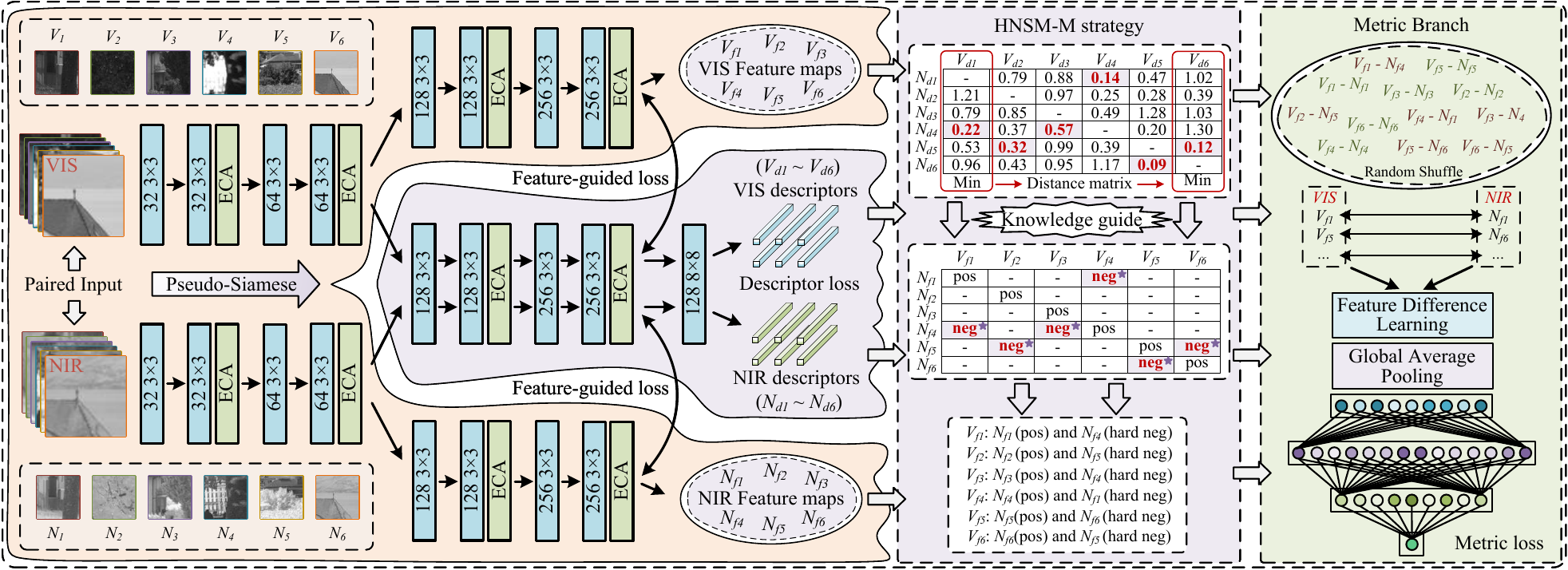}
    \vspace{-15pt}
    \caption{\hlA{The overall structure of our KGL-Net. Each blue block denotes a convolution operation, a Filter Response Normalization (FRN)~\cite{singh2020filter}, and a Thresholded Linear Unit (TLU)~\cite{singh2020filter}, where ``$n$\,\, $b \times b$'' means that the number of channels of the output feature map is $n$ and the convolution kernel size is $b \times b$. ``$ECA$'' denotes the efficient channel attention module~\cite{wang2020eca}}.}
    \label{fig:fig02}
    \vspace{-3pt}
\end{figure*}

\subsection{Descriptor Learning Methods}

Descriptor learning methods~\cite{simo2015discriminative,balntas2016learning,balntas2016pn,kumar2016learning,aguilera2017cross,zhang2017learning,mishchuk2017working,wang2017train,zheng2019hardness,wang2019better,tian2017l2,tian2019sosnet,tian2020hynet} use neural networks to extract high-level image patch features and measure their similarity by feature distance. The goal of this method is to make the feature distance between matching samples smaller and the feature distance between non-matching samples larger. \hlB{Existing research mainly focuses on optimizing loss functions}~\cite{simo2015discriminative,balntas2016learning,balntas2016pn,kumar2016learning,aguilera2017cross,zhang2017learning,wang2019better,tian2017l2,tian2019sosnet,tian2020hynet} \hlB{and sampling strategies}~\cite{mishchuk2017working,wang2017train,zheng2019hardness,tian2017l2}. In terms of the loss functions, hinge embedding loss~\cite{simo2015discriminative} and triplet loss~\cite{balntas2016learning,balntas2016pn} are commonly used. Unlike hinge embedding loss, which uses absolute distance, triplet loss is optimized based on relative distance, which is more stable. Therefore, subsequent research focuses on triplet loss and its variants~\cite{balntas2016learning,balntas2016pn,kumar2016learning,aguilera2017cross,zhang2017learning,wang2019better,tian2017l2,tian2019sosnet,tian2020hynet}. For example, Aguilera \etal~\cite{aguilera2017cross} propose a Q-Net loss suitable for four-output networks based on the triplet loss. In terms of the sampling strategies, the early L2-Net~\cite{tian2017l2} proposes a progressive sampling method that can generate many negative samples. However, these negative samples are mostly simple samples, which are difficult to provide effective feedback for network training. Therefore, subsequent research gradually turned to the design of effective hard negative sample mining strategies~\cite{mishchuk2017working,wang2017train,zheng2019hardness}. For example, HardNet~\cite{mishchuk2017working} selects the most challenging negative samples for training during mini-batch processing. Zheng \etal~\cite{zheng2019hardness} make full use of the information of all sample pairs by synthesizing hard negative samples. \hlB{In addition, some studies also focus on improving feature extractors~\cite{passov2026multi}. However, due to its structural limitations, descriptor learning mainly evaluates the relative similarity between image patches by directly computing feature distances, making it less effective in modeling patch-wise relations~\cite{yu2023feature,quan2019afd,quan2021element,yu2022multibranch,yu2023efficient}. Therefore, descriptor learning is generally fast and lightweight, but often limited in cross-spectral matching performance.}

\subsection{Metric Learning Methods}
Metric learning methods~\cite{yu2023feature,aguilera2016learning,quan2019afd,quan2021element,yu2022multibranch,yu2023efficient,han2015matchnet,zagoruyko2015learning,quan2021multi,quan2018cross,en2018ts,ko2023spectral} extract and measure image patch features in sequence by using neural networks, which have flexible feature relation modeling capabilities. In this type of method, the data is converted into multiple feature representations in a learnable feature space, and a nonlinear similarity measurement function is constructed for it to predict the matching probability between image patches. This method focuses on the design of the network structure. MatchNet~\cite{han2015matchnet} is one of the earliest metric learning methods. Subsequently, Zagoruyko \etal~\cite{zagoruyko2015learning} show that a 2-channel network that directly calculates feature similarity based on a fused feature space has better matching performance than the Siamese network and Pseudo-Siamese network. Inspired by this, Quan \etal~\cite{quan2021multi} consider the relation learning between image patch features is better than the learning of individual image patch features. Therefore, a multi-relation attention learning network (MRAN)~\cite{quan2021multi} is constructed and achieves excellent results. However, the above metric learning methods are designed for single-spectral image patch matching. Cross-spectral image patch matching is more complex and its research is relatively late. Inspired by~\cite{zagoruyko2015learning}, Aguilera \etal~\cite{aguilera2016learning} propose the Siamese network, the Pseudo-Siamese network and the 2-channel network for cross-spectral image patch matching. Subsequently, SCFDM~\cite{quan2018cross} extracts semantically invariant features of cross-spectral image patches in a shared semantic feature space and achieves significant performance improvement, which further reveals that the relation learning between image patch features is better than the learning of individual image patch features. Therefore, \hlB{recent cross-spectral image patch matching methods}~\cite{yu2023feature,quan2019afd,quan2021element,yu2022multibranch,yu2023efficient,en2018ts,ko2023spectral,zhang2025mssi} \hlB{focus on the extraction of richer and more diverse relations. However, simply improving feature extraction networks or relation mining modules for more comprehensive relation modeling has gradually shown performance bottlenecks.  To address this challenge, we make the first attempt to explore a stable and efficient bridge between descriptor learning and metric learning, aiming to establish a new knowledge-guided combined learning paradigm and achieve further performance breakthroughs.}

\begin{figure*}[htbp]
    \captionsetup{skip=5pt}
    \centering
    \includegraphics[width=\textwidth]{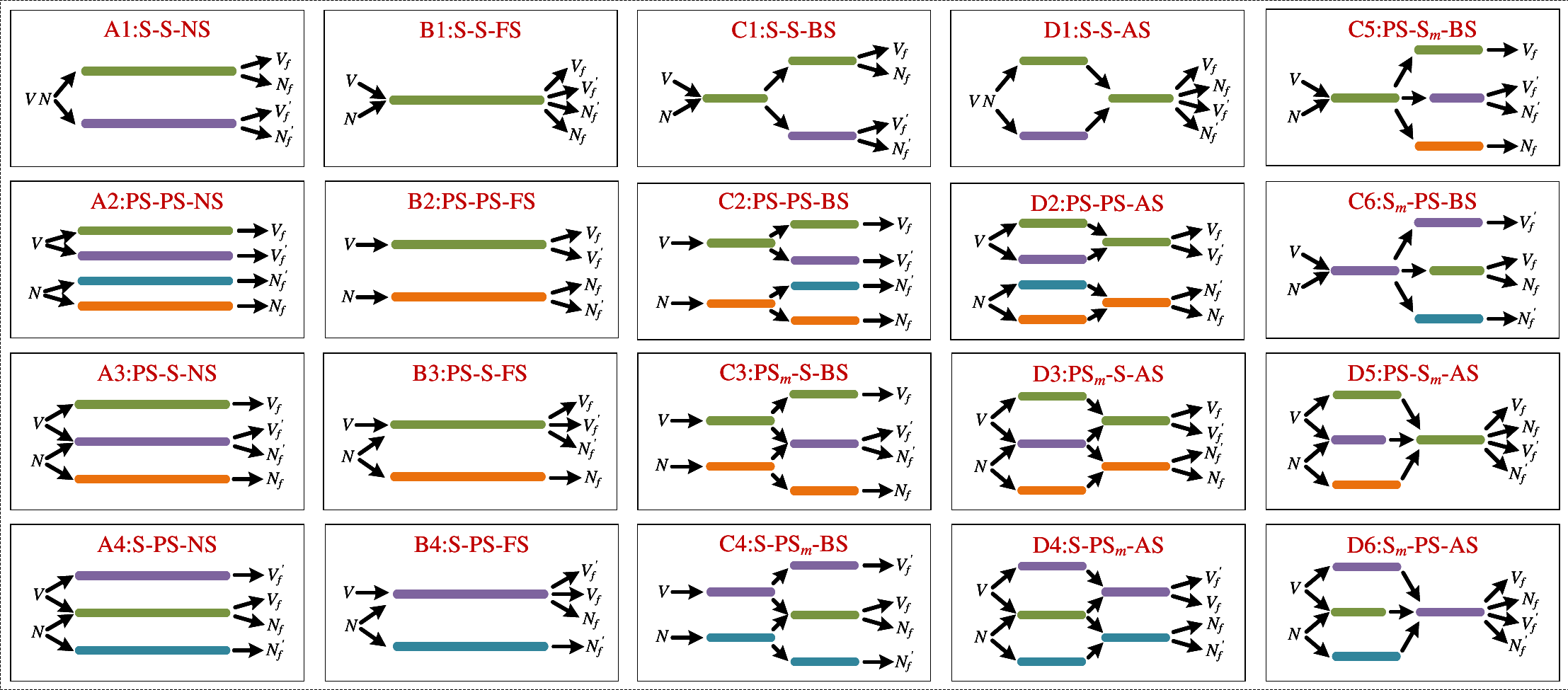}
    \vspace{-15pt}
    \caption{Various combined network architectures of descriptor learning and metric learning. In the subtitle, the three parts divided by short dashes denote the metric network structure, the descriptor network structure, and \hlB{parameter sharing between the two}. \textit{``S''} denotes Siamese. \textit{``P''} denotes \hlB{Pseudo-Siamese. \textit{``NS''}} denotes no parameters are shared. \textit{``FS''} denotes full parameter sharing. \textit{``BS''} denotes the parameters sharing of the front network layers. \textit{``AS''} denotes the parameters sharing of the later network layers. In addition, the presence of $m$ denotes that it is a dominant branch, which makes the structure of the corresponding part of another network consistent with it.}
    \label{fig:fig03}
    \vspace{-3pt}
\end{figure*}

\vspace{10pt}
\section{Methods}
\label{sec:methods}

In this section, we will first elaborate our motivations to explain the \textbf{``Why''}. Then, we will elaborate on our proposed KGL-Net to explain \textbf{``How''}. The overall structure of KGL-Net is shown in \cref{fig:fig02}.

\subsection{Motivation}
\label{sec:motivation}
\textbf{\textit{1) Performance bottleneck.}} For the cross-spectral image patch matching task, existing methods focus on improving the feature extraction network part to mine more sufficient and comprehensive feature relations between image patches~\cite{yu2023feature,quan2019afd,quan2021element,yu2022multibranch,yu2023efficient,quan2018cross,en2018ts,ko2023spectral}. However, as the network becomes more complex, the improvement in matching results gradually levels off. From \cref{fig:fig01}, compared with the results of MFD-Net, the FIL-Net only decreases the FPR95 by 3.4\% (from 0.88 to 0.85). At the same time, the performance bottlenecks of existing methods become increasingly apparent. \hlB{Considering} the advantages of descriptor learning and metric learning, we make the first attempt to build a stable and efficient bridge between descriptor learning and metric learning to overcome this bottleneck without increasing resource consumption.

\textbf{\textit{2) Feature extraction consistency.}} \hlA{There is feature extraction consistency between metric learning based on feature difference learning~\cite{quan2019afd} and descriptor learning based on Euclidean distance~\cite{quan2021element}.} They all focus on extracting the differential features of the image patch itself relative to other image patches. \hlA{Therefore, their gradient driving forces on the feature extractor are aligned, forming difference-driven consistency at the semantic level.} Taking a visible spectrum (VIS) image patch \textit{$V_p$} and a near infrared (NIR) image patch \textit{$N_p$} \hlB{as an example}, they are both passed through the feature extraction part ${\varphi_{mf}}$ of the metric network and the feature extraction part ${\varphi_{df}}$ of the descriptor network, respectively:
\vspace{-2pt}
\begin{gather}
({m_V},{m_N}) = {\varphi _{mf}}({V_p},{N_p}) \\
({d_V},{d_N}) = {\varphi _{df}}({V_p},{N_p})
\end{gather}

Then, the extracted features are operated via the metric branch based on feature difference learning and the descriptor processing based on Euclidean distance:
\vspace{-2pt}
\begin{gather}
{S_{out}} = {\varphi '_m}\left| {{m_V} - {m_N}} \right| \\
dis{t_{out}} = {\left\| {{{\hat d}_V} - {{\hat d}_N}} \right\|_2} = {\left\| {{{\varphi '}_d} \cdot {d_V} - {{\varphi '}_d} \cdot {d_N}} \right\|_2}
\end{gather}
where $S_{out}$ is the similarity output by the metric network. $dist_{out}$ is the distance output by the descriptor network. $\varphi '_m$ denotes the metric branch network part. $\varphi '_d$ denotes the network part in the descriptor network that converts high-level features into descriptors. ${\hat d_V}$ and ${\hat d_N}$ denote the VIS descriptor and NIR descriptor, respectively.

Feature difference learning and Euclidean distance \hlB{both essentially measure feature differences} and have similarities. They all drive the feature extraction network to extract the differential features of the image patch itself relative to other image patches. This provides a foundation for building a stable and efficient bridge between descriptor learning and metric learning. \hlA{Notably, feature extraction consistency is not entirely equivalent to feature consistency. Instead, it emphasizes that both components drive the feature extractor to reinforce the differences between cross-spectral image patches while suppressing redundant background information.}

\textbf{\textit{3) Feature mapping capabilities of the metric branch.}} We find that existing metric learning methods often ignore the feature mapping capability of the metric branch. The metric branch of the metric network is essentially equivalent to a learnable discriminator. We think that if we can provide more effective samples for learning, it will theoretically have stronger discriminative performance. However, this is difficult to achieve when using the input data form of existing metric learning methods, which use uniform and fixed positive and negative sample pairs. Therefore, we break the limitation, only use positive sample pairs in the dataset as inputs and randomly generate negative sample pairs for learning. However, directly generating negative samples often yields mostly simple examples. In addition, unlike the descriptor learning method, the metric branch of the metric learning method combines the features of the two branches and finally outputs a similarity, which results in the inability to directly perform effective hard negative sample mining. Therefore, we attempt to combine descriptor learning to achieve effective knowledge guidance for metric learning, thereby achieving effective hard negative sample mining for metric learning methods.

\renewcommand{\algorithmicrequire}{\textbf{Input:}}
\renewcommand{\algorithmicensure}{\textbf{Output:}}
\renewcommand{\thealgorithm}{1}
\begin{algorithm}[t]
\caption{KGL-Net Training (Batch image pairs input)}
\label{alg:kglnet}
\begin{algorithmic}[1] 
\Require Batch image pairs $\left\{ {({V_i},{N_{\rm{i}}})} \right\}_{i = 1}^{{N_b}}$; Descriptor network ${\varphi _d}$ (Feature extraction network part ${\varphi _{df}}$ + Descriptor conversion network part ${\varphi '_d}$); \hlB{Metric network} ${\varphi _m}$ (Feature extraction network part ${\varphi _{mf}}$ + Metrics branch ${\varphi '_m}$); Loss weights ($\alpha$, $\beta$); Learning rate $lr$.
\Ensure Updated metric network parameters ${\Theta _m}$ and descriptor network parameters ${\Theta _d}$.

\State \textbf{Initialize:} Give the initial value to ${\varphi _d}$, ${\varphi _m}$, $\alpha$, $\beta$, $lr$.

\State \textbf{Step 1}: Input batch image pairs $\left\{ {({V_i},{N_{\rm{i}}})} \right\}_{i = 1}^{{N_b}}$ into the feature extraction parts (${\varphi _{df}}$, ${\varphi _{mf}}$) of descriptor network and metric network to obtain high-level feature maps:
\[\left\{ {(d_V^i,d_N^i)} \right\}_{i = 1}^{{N_b}} \leftarrow {\varphi _{df}}\left\{ {({V_i},{N_i})} \right\}_{i = 1}^{{N_b}}\]
\[
\left\{ {(m_V^i,m_N^i)} \right\}_{i = 1}^{{N_b}} \leftarrow {\varphi _{mf}}\left\{ {({V_i},{N_i})} \right\}_{i = 1}^{{N_b}}
\]

\State \textbf{Step 2}: Use $\left\{ {(d_V^i,d_N^i)} \right\}_{i = 1}^{{N_b}}$ to construct a batch feature matrix:
\[
{M_{ij}} = {\left\| {{{\varphi '}_d}(d_V^i) - {{\varphi '}_d}(d_N^j)} \right\|_2}\quad i,j \in [1,{N_b}]
\]

\State \textbf{Step 3}: For each sample, according to ${M_{ij}}$ to construct hard negative sample pairs:
\[(m_V^i,m_N^{Inde{x_i}}) \leftarrow Inde{x_i} = \mathop {\arg \min }\limits_{i \ne j} {M_{ij}}\]

\State \textbf{Step 4}: The predicted matching scores of positive and negative samples are calculated by the metric branch ${\varphi '_m}$:
\[S_{pos}^i = {\varphi '_m}(m_V^i,m_N^i)\]
\[S_{neg}^i = {\varphi '_m}(m_V^i,m_N^{Inde{x_i}})\]

\State \textbf{Step 5}: Based on the predicted matching scores, calculate the binary cross-entropy loss ${L_m}$. Based on ${M_{ij}}$, the triplet margin loss~\cite{tian2020hynet} is used to calculate the descriptor loss ${L_d}$. Based on the obtained high-level features ($d_V^i$, $d_N^i$, $m_V^i$, $m_N^i$), the feature-guided losses ($L_{fg}^V$, $L_{fg}^N$) are calculated according to \Cref{eq:lfgv} and \Cref{eq:lfgn}.

\State \textbf{Step 6}: \hlB{Sum all sub-loss} to get the total loss $L$:
\[L = {L_d} + \alpha {L_m} + \beta (L_{fg}^V + L_{fg}^N)\]

\State \textbf{Step 7}: Based on the Adam optimizer, update the network parameters (${\Theta _d}$, ${\Theta _m}$):
\[{\Theta _d} \leftarrow {\Theta _d} - lr{\nabla _{{\Theta _d}}}L\]
\[{\Theta _m} \leftarrow {\Theta _m} - lr{\nabla _{{\Theta _m}}}L\]

\State \textbf{Step 8}: Repeat Step 1-7 until the training is completed.

\State \textbf{Return:} ${\Theta _m}$, ${\Theta _d}$.

\end{algorithmic} 
\end{algorithm}

\subsection{Overview}
\label{sec:overview}
To break the performance bottleneck of current research, we break away from the inertial thinking of building a more complex feature relation extraction network structure. From \cref{fig:fig02}, we break the input format limitation of existing metric learning methods and construct an end-to-end knowledge-guided learning network (KGL-Net), which is the first to realize effective hard negative sample mining for metric learning and establish a stable and efficient bridge between descriptor learning and metric learning. Our KGL-Net has achieved \hlA{significant} performance advantages while abandoning complex network structures, and it has strong scalability. Specifically, KGL-Net can be divided into three parts: \hlB{feature extraction network, hard negative sample mining and metric branch.} For the feature extraction network part, it includes the feature extraction part of the metric network (light yellow area) and the descriptor network (light purple area). \hlB{The low-level network layers in the descriptor network are shared with the metric network.} In the hard negative sample mining part, a HNSM-M strategy is proposed and used, which uses the descriptors output by the descriptor network to obtain the position index of hard negative samples and guides the features extracted by the metric network. For the metric branch, we apply the feature difference learning~\cite{yu2023feature,quan2019afd} and three fully connected layers with 512, 256, and 1 neurons, which can effectively extract the difference features between image patches and perform feature mapping to determine whether the input image patch pairs are matched.

\subsection{Combined Network Architecture}
\label{sec:combined network architecture}
To ensure the stability and efficiency of the bridge between descriptor learning and metric learning, we systematically explore 20 combined network architectures as shown in \cref{fig:fig03}, including no parameters shared (A1-A4), full parameter sharing (B1-B4), parameter sharing of low-level network layers (C1-C6), and parameter sharing of high-level network layers (D1-D6). The C3 architecture is finally selected. In C3, the metric network is a Pseudo-Siamese structure, the descriptor network is a Siamese structure, and the low-level network layers of the descriptor network become a Pseudo-Siamese structure to achieve the sharing of the low-level network layers of the metric network.

Specifically, paired image patches are first input into two feature extraction branches that do not share parameters to extract their respective low-level detail features. This part has two characteristics. On the one hand, the parameters of different spectral branches are not shared, which is conducive to fully extracting the detailed features of the different spectral image patches. On the other hand, the parameters of the descriptor network and the metric network are shared, which not only helps both to effectively guide the extraction of low-level features, but also provides a basis for subsequent effective knowledge guidance of hard negative samples. Then, the low-level feature maps of the two branches are input into the Pseudo-Siamese metric network, which helps to further fully extract the high-level semantic features of each \hlB{spectral modality}~\cite{yu2022multibranch}. Meanwhile, the low-level feature maps are also input into the Siamese descriptor network to extract high-level semantic features that tend to be in the same feature space~\cite{quan2022self}, which helps to achieve more refined hard negative sample mining, thereby providing more effective knowledge guidance. \hlB{Overall, C3 better balances low-level sharing and high-level specialization, using shared low-level layers for aligned detail representations while preserving the strengths of descriptor and metric branches in descriptor-space distance ranking and cross-spectral discriminative matching, respectively.}

\subsection{HNSM-M strategy}
\label{sec:hnsm-m strategy}
Different from the existing methods, we think that the feature mapping ability of the metric branch should receive more attention. To improve the discriminative ability of the metric branch with the same structure, a HNSM-M strategy is proposed. From \cref{fig:fig02}, for the input positive sample pairs, we use the descriptor network to generate descriptors and use the metric network based on feature difference learning to generate high-level feature maps. Benefiting from the sharing of low-level network layers, the constraints of feature-guided loss (\cref{sec:loss function}), and the feature extraction consistency, there is a corresponding relation between the descriptors and the high-level feature maps, and they can guide each other's learning.

Taking VIS-NIR as an example, for a training batch containing $N_b$ sample pairs, the corresponding positive descriptor pairs $\{ \hat d_V^i,\hat d_N^i\} _{i = 1 \ldots N_b}$ and the corresponding positive high-level feature map pairs $\{ m_V^i,m_N^i\} _{i = 1 \ldots N_b}$ are obtained after passing through the feature extraction part of the descriptor network and the metric network. First, we use the obtained descriptors to construct the feature distance matrix to traverse and obtain the feature distances of the positive sample pairs and all negative sample pairs:
\begin{gather}
{M_{ij}} = {\left\| {\hat d_V^i - \hat d_N^j} \right\|_2}
\end{gather}
where $M_{ij}$ denotes the feature distance matrix. \hlA{Its calculations are performed using vectorized GPU matrix operations, which are low in complexity and fast in speed.}

Second, based on the $M_{ij}$, we can efficiently locate the NIR descriptor closest to a VIS descriptor in a batch:
\begin{gather}
Inde{x_i} = \mathop {argmin}\limits_{i \ne j} {M_{ij}}
\end{gather}

Finally, the index helps the VIS feature map locate its corresponding NIR feature map in the metric network, forming a hard negative sample pair.

By randomly shuffling and sampling an equal number of positive and hard negative sample pairs as input into the metric branch, the metric branch can have better discrimination ability and also promote the active optimization of the feature extraction network. \hlA{At the same time, the HNSM-M strategy performs the sorting step without gradient backpropagation, uses fully vectorized computation, and contains no trainable parameters. The additional time and memory overhead it introduces is minimal. In addition, it is executed only during the training phase and thus does not affect the inference time. According to statistics, on the VIS-NIR patch dataset, the training time per epoch for KGL-Net with and without the HNSM-M strategy is 0.205 h and 0.204 h, respectively. This verifies the efficiency of the HNSM-M strategy.}

\begin{figure*}[]
    \captionsetup{skip=5pt}
    \centering
    \includegraphics[width=\textwidth]{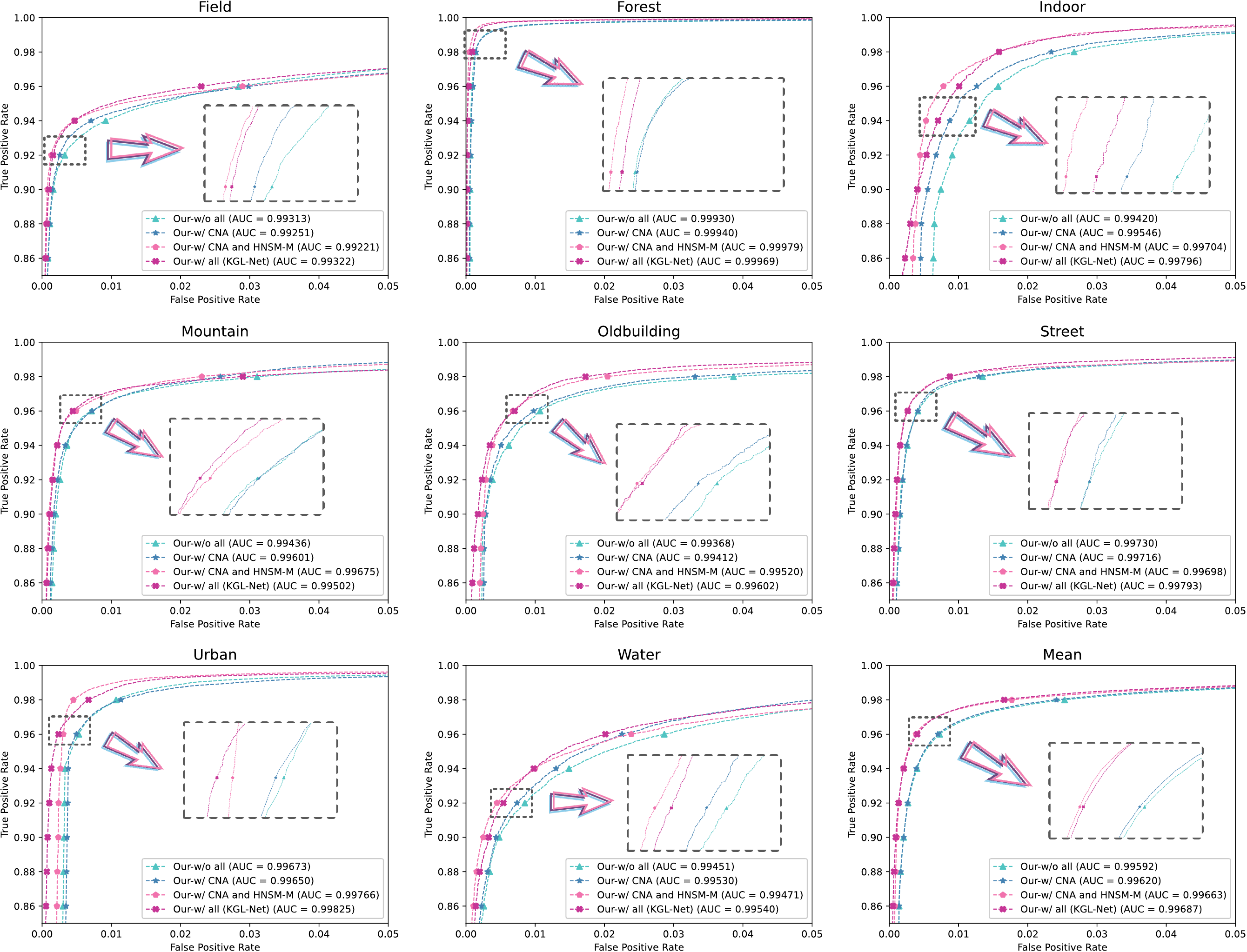}
    \vspace{-15pt}
    \caption{\hlA{Comparison of the receiver operating characteristic (ROC) curve and area under the curve (AUC) of each scheme in the break-down ablation experiment. From ``Mean'', compared with other schemes, \textit{``Our-w/ all (KGL-Net)''} has the highest AUC. From each subset, \textit{``Our-w/ all (KGL-Net)''} is also mostly at the top and has the highest AUC.}}
    \label{fig:fig07}
    \vspace{0pt}
\end{figure*}

\begin{figure*}[]
    \captionsetup{skip=5pt}
    \centering
    \includegraphics[width=\textwidth]{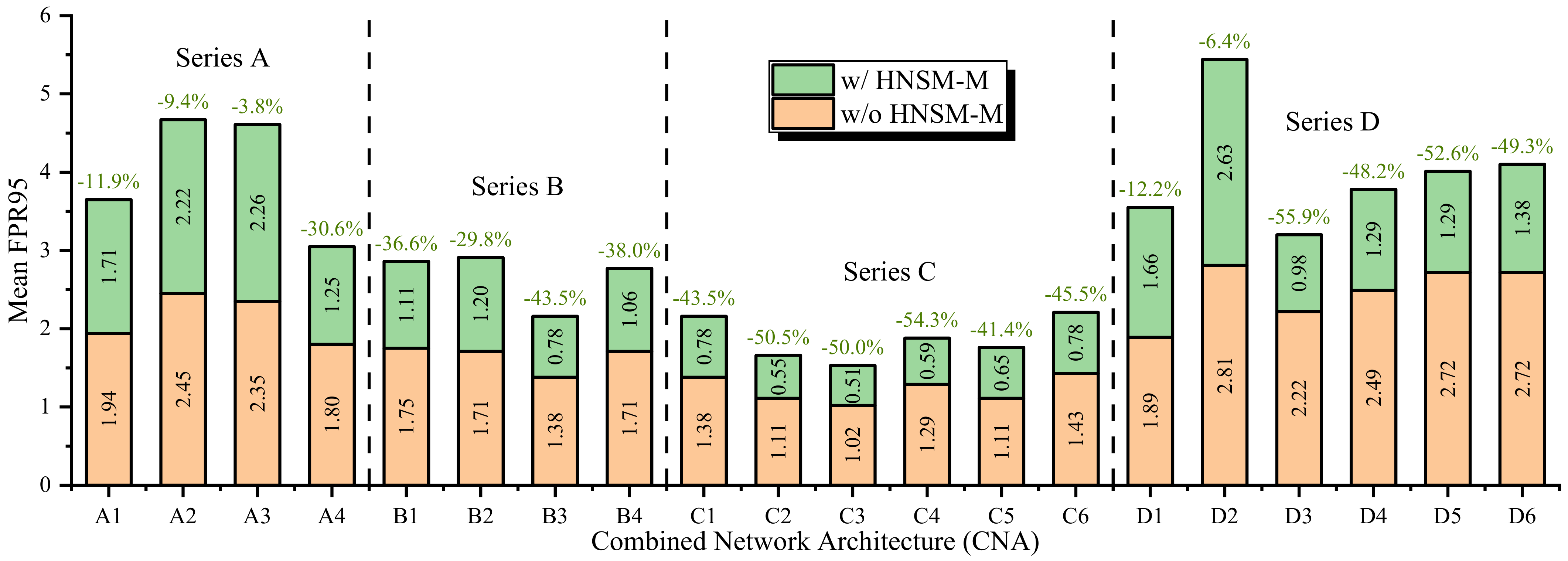}
    \vspace{-15pt}
    \caption{Comparison of different combined network architectures on the VIS-LWIR patch dataset. The percentages denote the reduction in FPR95 achieved by applying the HNSM-M strategy compared to the baseline without HNSM-M strategy. On one hand, Series C exhibits significantly lower FPR95 compared to other series architectures. On the other hand, Using the HNSM-M strategy, the FPR95 decreased by 3.8\% (from 2.35 to 2.26) - 55.9\% (from 2.22 to 0.98).}
    \label{fig:fig08}
    \vspace{-4pt}
\end{figure*}

\subsection{Loss Function}
\label{sec:loss function}
This work aims to establish a stable and efficient bridge between descriptor learning and metric learning. \hlB{We find the feature extraction consistency between the two methods.} However, owing to their inherent structural characteristics, although the high-level feature maps reflect semantic features, the output of the descriptor network tends to reflect global features more, whereas the output of the metric network tends to reflect local features more. To ensure that the hard negative sample positions obtained by the descriptor can have strict guidance for the metric network, we construct a feature-guided loss to achieve effective mutual guidance of the high-level feature maps of the two methods: 
\begin{gather}
L_{fg}^V = \frac{1}{{{N_b}}}\sum\limits_{i = 1}^{{N_b}} {{{\left\| {m_V^i - d_V^i} \right\|}_2}}  \label{eq:lfgv} \\
L_{fg}^N = \frac{1}{{{N_b}}}\sum\limits_{i = 1}^{{N_b}} {{{\left\| {m_N^i - d_N^i} \right\|}_2}} \label{eq:lfgn}
\end{gather}
where $L_{fg}^V$ and $L_{fg}^N$ denote the feature-guided loss outputs of the VIS branch and the NIR branch, respectively. 

In addition to the feature-guided loss, we employ descriptor loss and metric loss to constrain the learning of the descriptor and metric network components in KGL-Net. The descriptor loss is a hybrid similarity measure for triplet margin loss~\cite{tian2020hynet}, whereas the metric loss uses the conventional binary cross-entropy loss~\cite{yu2023feature}. The total loss is calculated as follows:
\begin{gather}
L = {L_d} + \alpha {L_m} + \beta (L_{fg}^V + L_{fg}^N)
\end{gather}
where $L$, $L_d$ and $L_m$ denote the total loss, descriptor loss and metric loss, respectively. \hlB{Details of the descriptor loss can be found in}~\cite{tian2020hynet}. Both $\alpha$ and $\beta$ are set to 1. The network optimization of our KGL-Net is shown in \Cref{alg:kglnet}.

\begin{table}[t]
\centering
\begin{minipage}[t]{\columnwidth}
    
    \caption{Break-down ablation experiments on the VIS-NIR patch dataset. $CNA$ denotes the combined network architecture. $FGL$ denotes feature-guided loss. ``$w/o \, all$'' refers to using only the metric network with a Pseudo-Siamese structure.}
    \vspace{-4pt}
    \label{tab:tab02}
    \setlength{\tabcolsep}{2pt}
    \renewcommand{\arraystretch}{1.2}
    \resizebox{\columnwidth}{!}{
    \begin{tabular}{c|ccc|c}
    \hline
                              & \multicolumn{3}{c|}{Variants} &                           \\
    \multirow{-2}{*}{Methods} & CNA     & HNSM-M     & FGL    & \multirow{-2}{*}{FPR95 ↓} \\ \hline
    Our-w/o all               & \ding{55}       & \ding{55}          & \ding{55}      & 0.88 (-38.6\%)            \\
    Our-w/ CNA                & \ding{51}       & \ding{55}          & \ding{55}      & 0.77 (-29.9\%)            \\
    Our-w/ CNA and HNSM-M     & \ding{51}       & \ding{51}          & \ding{55}      & 0.57 (-5.3\%)             \\ \hline
    \rowcolor[HTML]{CFCECE} 
    KGL-Net (Ours)            & \ding{51}       & \ding{51}          & \ding{51}      & \textbf{0.54}             \\ \hline
    \end{tabular}
    }
\end{minipage}

\hspace{0.05\textwidth} 

\begin{minipage}[t]{\columnwidth}
    \caption{Comparison of different combined network architectures on the VIS-NIR patch dataset. The results with background color denote the results obtained via the HNSM-M strategy. \hlA{The percentages in parentheses denote the relative change.}}
    \vspace{-4pt}
    \label{tab:tab03}
    \renewcommand{\arraystretch}{1.2}
    \setlength{\tabcolsep}{1.5pt}
    \resizebox{\columnwidth}{!}{
    \begin{tabular}{c|c|c|c|c|c|c|c}
    \hline
    CNA                  & FPR95 ↓                               & \multicolumn{1}{l|}{CNA} & FPR95 ↓                               & \multicolumn{1}{l|}{CNA} & FPR95 ↓                               & CNA                  & FPR95 ↓                               \\ \hline
                         & 1.09 (-33.0\%)                        &                          & 0.98 (-6.1\%)                         &                          & 0.84 (-20.2\%)                        &                      & 0.96 (-26.0\%)                        \\ \cline{2-2} \cline{4-4} \cline{6-6} \cline{8-8} 
    \multirow{-2}{*}{A1} & \cellcolor[HTML]{CFCECE}\textbf{0.73} & \multirow{-2}{*}{A2}     & \cellcolor[HTML]{CFCECE}\textbf{0.92} & \multirow{-2}{*}{A3}     & \cellcolor[HTML]{CFCECE}\textbf{0.67} & \multirow{-2}{*}{A4} & \cellcolor[HTML]{CFCECE}\textbf{0.71} \\ \hline
                         & 1.07 (-6.5\%)                         &                          & 1.44 (-20.1\%)                        &                          & 1.02 (-19.6\%)                        &                      & 1.22 (-16.4\%)                        \\ \cline{2-2} \cline{4-4} \cline{6-6} \cline{8-8} 
    \multirow{-2}{*}{B1} & \cellcolor[HTML]{CFCECE}\textbf{1.00} & \multirow{-2}{*}{B2}     & \cellcolor[HTML]{CFCECE}\textbf{1.15} & \multirow{-2}{*}{B3}     & \cellcolor[HTML]{CFCECE}\textbf{0.82} & \multirow{-2}{*}{B4} & \cellcolor[HTML]{CFCECE}\textbf{1.02} \\ \hline
                         & 0.93 (-35.5\%)                        &                          & 0.98 (-37.8\%)                        &                          & 0.81 (-33.3\%)                        &                      & 1.02 (-37.3\%)                        \\ \cline{2-2} \cline{4-4} \cline{6-6} \cline{8-8} 
    \multirow{-2}{*}{C1} & \cellcolor[HTML]{CFCECE}\textbf{0.60} & \multirow{-2}{*}{C2}     & \cellcolor[HTML]{CFCECE}\textbf{0.61} & \multirow{-2}{*}{C3}     & \cellcolor[HTML]{CFCECE}\textbf{0.54} & \multirow{-2}{*}{C4} & \cellcolor[HTML]{CFCECE}\textbf{0.64} \\ \hline
                         & 1.49 (-35.6\%)                        &                          & 1.75 (-38.9\%)                        &                          & 1.52 (-39.5\%)                        &                      & 1.34 (-17.9\%)                        \\ \cline{2-2} \cline{4-4} \cline{6-6} \cline{8-8} 
    \multirow{-2}{*}{D1} & \cellcolor[HTML]{CFCECE}\textbf{0.96} & \multirow{-2}{*}{D2}     & \cellcolor[HTML]{CFCECE}\textbf{1.07} & \multirow{-2}{*}{D3}     & \cellcolor[HTML]{CFCECE}\textbf{0.92} & \multirow{-2}{*}{D4} & \cellcolor[HTML]{CFCECE}\textbf{1.10} \\ \hline
                         & 0.88 (-31.8\%)                        &                          & 0.97 (-25.8\%)                        &                          & 1.24 (-21.0\%)                        &                      & 1.28 (-30.5\%)                        \\ \cline{2-2} \cline{4-4} \cline{6-6} \cline{8-8} 
    \multirow{-2}{*}{C5} & \cellcolor[HTML]{CFCECE}\textbf{0.60} & \multirow{-2}{*}{C6}     & \cellcolor[HTML]{CFCECE}\textbf{0.72} & \multirow{-2}{*}{D5}     & \cellcolor[HTML]{CFCECE}\textbf{0.98} & \multirow{-2}{*}{D6} & \cellcolor[HTML]{CFCECE}\textbf{0.89} \\ \hline
    \end{tabular}
    }
\end{minipage}

\hspace{0.05\textwidth} 

\begin{minipage}[t]{\columnwidth}
    \caption{\hlA{Comparison of different network layer sharing based on the C3 architecture on the VIS-NIR patch dataset and VIS-LWIR patch dataset. In the FPR95 results presentation, the left and right sides of the diagonal line respectively show the results for VIS-NIR and VIS-LWIR. Sharing layer addition rules are added sequentially from the low to the high.}}
    \vspace{-4pt}
    \label{tab:tab04}
    \renewcommand{\arraystretch}{1.2}
    \setlength{\tabcolsep}{6pt}
    \resizebox{\columnwidth}{!}{
    \begin{tabular}{c|c|c|c|c|c}
    \hline
    Layers & FPR95 ↓ & Layers             & FPR95 ↓                               & Layers & FPR95 ↓ \\ \hline
    0             & 0.67 / 2.26    & 3                         & 0.54 / 0.88                                  & 6             & 0.65 / 0.51    \\
    1             & 0.59 / 1.85    & \cellcolor[HTML]{CFCECE}4 & \cellcolor[HTML]{CFCECE}\textbf{0.54 / 0.51} & 7             & 0.75 / 0.53    \\
    2             & 0.58 / 1.52    & 5                         & 0.59 / 0.51                                  & 8             & 0.93 / 0.74    \\ \hline
    \end{tabular}
    }

\end{minipage}

\hspace{0.05\textwidth} 

\begin{minipage}[t]{\columnwidth}
\caption{\hlA{Exploration of different projection head settings before feature-guided loss on the VIS-NIR patch dataset. ``512-512-512'' denotes 3 hidden layers with 512 neurons each.}}
\vspace{-4pt}
\label{tab:tabs-projection-head}
\setlength{\tabcolsep}{8pt}
\renewcommand{\arraystretch}{1.2}
\resizebox{\columnwidth}{!}{
\begin{tabular}{c|c|c}
\hline
Projection head & FPR95 & Training time per epoch (h) \\ \hline
{\cellcolor[HTML]{CFCECE} N/A}               & {\cellcolor[HTML]{CFCECE} \textbf{0.54}}   & {\cellcolor[HTML]{CFCECE} \textbf{0.205}}                       \\ \hline
512             & 0.55 (-1.8\%)  & 0.206 (-0.5\%)         \\ \hline
512-512         & \textbf{0.54} (-0.0\%)   & 0.208 (-1.4\%)         \\ \hline
512-512-512     & \textbf{0.54} (-0.0\%)   & 0.212 (-3.3\%)          \\ \hline
\end{tabular}
}
\end{minipage}
\vspace{-8pt}
\end{table}

\vspace{10pt}
\section{Experiments}
\label{sec:experiments}

\subsection{Datasets}
\label{sec:datasets}
\hlA{Six public multimodal datasets are used in our experiments, including the VIS-NIR patch dataset}~\cite{aguilera2016learning}, \hlA{VIS-LWIR patch dataset}~\cite{yu2023feature}, \hlA{LLVIP patch dataset}~\cite{yu2025relational}, \hlA{OS patch dataset}~\cite{yu2023efficient}, \hlA{RIRE patch dataset}~\cite{yu2025relational}, \hlA{and VIS-NIR registration dataset}~\cite{yu2025relational}. \hlA{Except for the VIS-NIR registration dataset, which is an image registration dataset, the other five are image patch matching datasets. The VIS-NIR patch dataset contains visible spectrum (VIS) and near-infrared (NIR). The VIS-LWIR patch dataset and LLVIP patch dataset contain VIS and long-wave infrared (LWIR). The OS patch dataset contains VIS and synthetic aperture radar (SAR). The RIRE patch dataset contains computed tomography (CT) and magnetic resonance imaging (MRI). All image patches are 64×64 pixels. More details can be found in the \hlB{\textit{Supplementary Material}}.}

\begin{figure}[t]
  \captionsetup{skip=0pt}
  \centering
   \includegraphics[width=0.7\columnwidth]{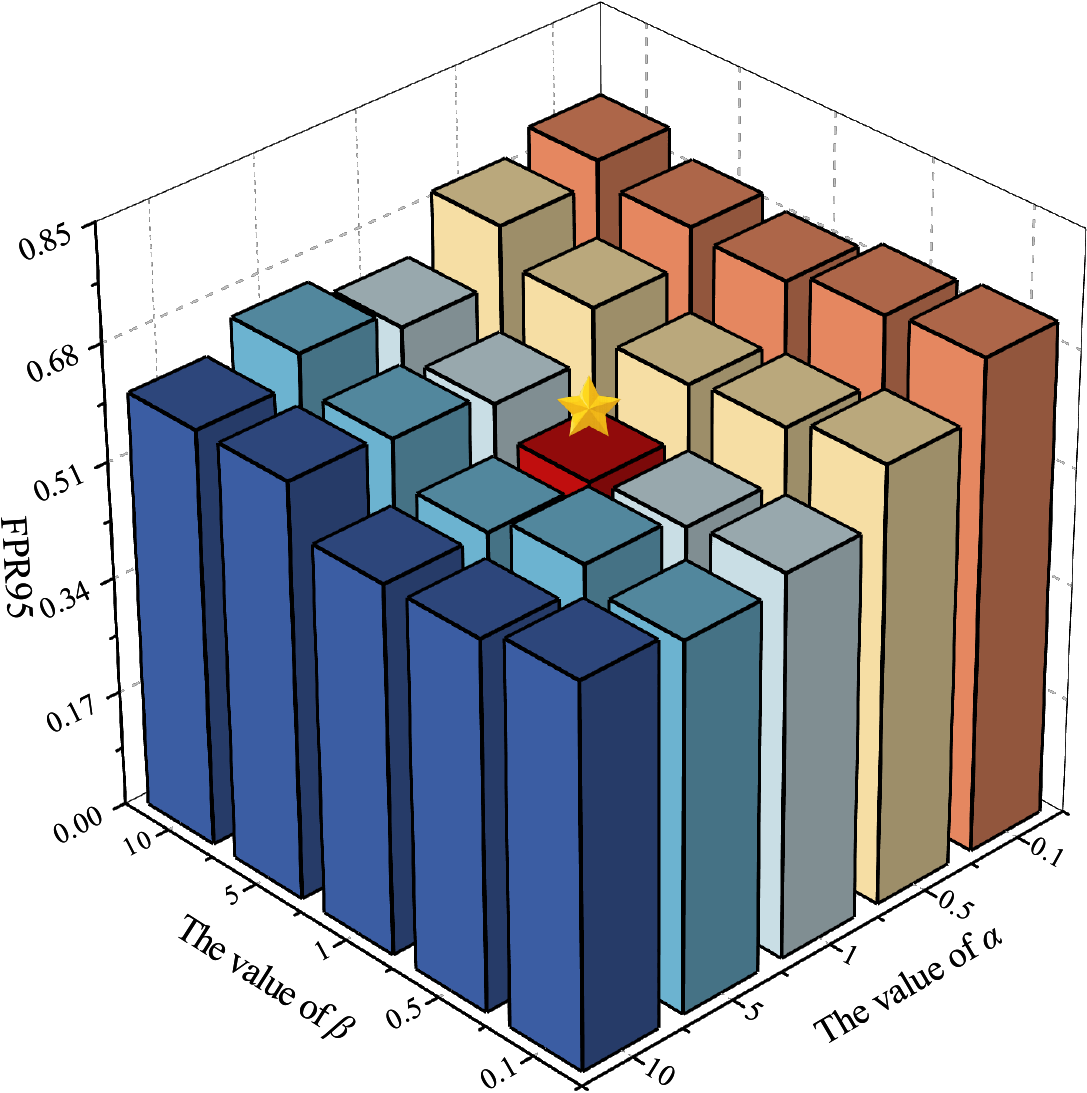}
   \caption{\hlA{Exploration of the $\alpha$ and $\beta$ in the feature-guided loss on the VIS-NIR patch dataset. The} \textcolor{darkred}{\hlA{red bar}} \hlA{denotes the optimal value, which is 0.54.}}
   \label{fig:figs-hyperparameters-fgl}
   \vspace{-5pt}
\end{figure}

\subsection{Experimental Details}
\label{sec:experimental details}
In the experiment, the batch size is 256, and the GPU is RTX 3090 24G. To ensure that the network can fully learn hard samples, the training epochs on the OS patch dataset is 200, and the training epochs on other datasets is 100. \hlA{The training time for KGL-Net on the VIS-NIR patch dataset, VIS-LWIR patch dataset, LLVIP patch dataset, OS patch dataset, and RIRE patch dataset are 0.205 h, 0.006 h, 0.067 h, 0.036 h, and 0.022 h per epoch, respectively.} Consistent with~\cite{mishchuk2017working,tian2017l2,tian2019sosnet,tian2020hynet}, the descriptor length is 128. In addition, instead of using a single learning rate, we use a learning rate of $5e^{\text{\scriptsize -3}}$ for the feature extraction part and $5e^{\text{\scriptsize -5}}$ for the metric branch and use the Adam~\cite{kingma2014adam} optimizer to train the model. For the evaluation metric, consistent with existing studies~\cite{yu2023feature,quan2022deep,quan2022self,aguilera2016learning,quan2019afd,quan2021element,yu2022multibranch,yu2023efficient,simo2015discriminative,balntas2016learning,balntas2016pn,kumar2016learning,aguilera2017cross,zhang2017learning,mishchuk2017working,wang2017train,zheng2019hardness,wang2019better,tian2017l2,tian2019sosnet,tian2020hynet,han2015matchnet,zagoruyko2015learning,quan2021multi,quan2018cross,en2018ts,ko2023spectral}, \hlA{the false positive rate at 95\% recall (FPR95) is used for the image patch matching. Lower FPR95 values indicate better performance. For image registration, the root mean square error (RMSE), the registered image $N_m$, and the image registration rate $I_{rr}$ are used. Registered images with an RMSE less than 5 are considered registered images.} In addition, \hlB{to} generate more samples to enhance the generalization of the network, we randomly process the images through data augmentation strategies such as flipping, rotation, and contrast adjustment. \hlA{The Gaussian blur range is [0, 1], the contrast adjustment range is [0.5, 1.5], the scale range is [0.95, 1.1], the translation range is [-0.05, 0.05], and the random cropping and filling range is [-0.1, 0.1]. Detailed settings can be found in our} \href{https://github.com/YuChuang1205/KGL-Net}{open-source project}. In the presented experimental results, \textbf{bold} denotes the optimal result, and \underline{underline} denotes the suboptimal result.

\begin{table*}[]
\caption{Comparison of FPR95 among KGL-Net and nineteen SOTA methods on the VIS-NIR patch dataset.}
\label{tab:tab01}
\setlength{\tabcolsep}{2.5mm}
\renewcommand{\arraystretch}{1.12}
\resizebox{\textwidth}{!}{
\begin{tabular}{cccccccccc}
\hline
Models             & Field         & Forest        & Indoor        & Mountain      & Oldbuilding   & Street        & Urban         & Water         & Mean          \\ \hline
\multicolumn{10}{c}{Traditional methods}                                                                                                                                                      \\
SIFT\textsuperscript{~\cite{lowe2004distinctive}}\textsuperscript{[IJCV]}                   & 39.44         & 11.39         & 10.13         & 28.63         & 19.69         & 31.14         & 10.85         & 40.33         & 23.95         \\
GISIFT\textsuperscript{~\cite{firmenichy2011multispectral}}\textsuperscript{[ICIP]}                 & 34.75         & 16.63         & 10.63         & 19.52         & 12.54         & 21.80         & 7.21          & 25.78         & 18.60         \\
LGHD\textsuperscript{~\cite{aguilera2015lghd}}\textsuperscript{[ICIP]}                   & 16.52         & 3.78          & 7.91          & 10.66         & 7.91          & 6.55          & 7.21          & 12.76         & 9.16          \\ \hline
\multicolumn{10}{c}{Descriptor learning}                                                                                                                                                      \\
PN-Net\textsuperscript{~\cite{balntas2016pn}}\textsuperscript{[arXiv]}                & 20.09         & 3.27          & 6.36          & 11.53         & 5.19          & 5.62          & 3.31          & 10.72         & 8.26          \\
Q-Net\textsuperscript{~\cite{aguilera2017cross}}\textsuperscript{[Sensors]}               & 17.01         & 2.70          & 6.16          & 9.61          & 4.61          & 3.99          & 2.83          & 8.44          & 6.91          \\
L2-Net\textsuperscript{~\cite{tian2017l2}}\textsuperscript{[CVPR]}                 & 13.67         & 2.48          & 4.63          & 8.87          & 4.12          & 5.58          & 1.54          & 6.55          & 5.93          \\
HardNet\textsuperscript{~\cite{mishchuk2017working}}\textsuperscript{[NeurIPS]}             & 5.61          & 0.15          & 1.50          & 3.14          & 1.10          & 1.93          & 0.69          & 2.29          & 2.05          \\
SOSNet\textsuperscript{~\cite{tian2019sosnet}}\textsuperscript{[CVPR]}                 & 5.94          & 0.13          & 1.53          & 2.45          & 0.99          & 2.04          & 0.78          & 1.90          & 1.97          \\
HyNet\textsuperscript{~\cite{tian2020hynet}}\textsuperscript{[NeurIPS]}               & 4.50          & 0.07          & 1.09          & 1.80          & 0.83          & 0.52          & 0.53          & 1.91          & 1.41          \\ \hline
\multicolumn{10}{c}{Metric learning}                                                                                                                                                          \\
Siamese\textsuperscript{~\cite{aguilera2016learning}}\textsuperscript{[CVPRW]}               & 15.79         & 10.76         & 11.60         & 11.15         & 5.27          & 7.51          & 4.60          & 10.21         & 9.61          \\
Pseudo-Siamese\textsuperscript{~\cite{aguilera2016learning}}\textsuperscript{[CVPRW]}        & 17.01         & 9.82          & 11.17         & 11.86         & 6.75          & 8.25          & 5.65          & 12.04         & 10.31         \\
2-channel\textsuperscript{~\cite{aguilera2016learning}}\textsuperscript{[CVPRW]}              & 9.96          & 0.12          & 4.40          & 8.89          & 2.30          & 2.18          & 1.58          & 6.40          & 4.47          \\
SCFDM\textsuperscript{~\cite{quan2018cross}}\textsuperscript{[ACCV]}                  & 7.91          & 0.87          & 3.93          & 5.07          & 2.27          & 2.22          & 0.85          & 4.75          & 3.48          \\
MR\_3A\textsuperscript{~\cite{quan2021multi}}\textsuperscript{[TIP]}                  & 4.21          & 0.11          & 1.12          & 0.87          & 0.67          & 0.56          & 0.43          & 1.90          & 1.23          \\
AFD-Net\textsuperscript{~\cite{quan2019afd,quan2021element}}\textsuperscript{[ICCV Oral, TNNLS]} & 3.47          & 0.08          & 1.48          & \underline{0.68}    & 0.71          & 0.42          & 0.29          & 1.48          & 1.08          \\
SPIMNet\textsuperscript{~\cite{ko2023spectral}}\textsuperscript{[Inf. Fusion]}         & \underline{2.28}    & 0.09          & 1.62          & 0.88          & 0.69          & 0.29          & 0.42          & 2.26          & 1.07          \\
EFR-Net\textsuperscript{~\cite{yu2023efficient}}\textsuperscript{[TGRS]}                & 2.84          & 0.07          & 1.09          & 0.79          & 0.61          & 0.50          & 0.34          & 1.65          & 0.99          \\
MFD-Net\textsuperscript{~\cite{yu2022multibranch}}\textsuperscript{[TGRS]}                & 2.59          & \textbf{0.02} & 1.24          & 0.95          & \underline{0.48}    & \underline{0.24}    & \underline{0.12}    & \underline{1.44}    & 0.88          \\
FIL-Net\textsuperscript{~\cite{yu2023feature}}\textsuperscript{[TIP]}                  & 2.61          & 0.04          & \underline{0.98}    & 0.73          & 0.49          & 0.34          & \textbf{0.10} & 1.49          & \underline{0.85}    \\
\rowcolor[HTML]{CFCECE} 
\textbf{KGL-Net (Ours)}                       & \textbf{1.00} & \underline{0.03}    & \textbf{0.83} & \textbf{0.28} & \textbf{0.47} & \textbf{0.20} & 0.17          & \textbf{1.37} & \textbf{0.54} \\ \hline
\end{tabular}
}
\end{table*}

\begin{figure*}[]
    \captionsetup{skip=5pt}
    \centering
    \includegraphics[width=\textwidth]{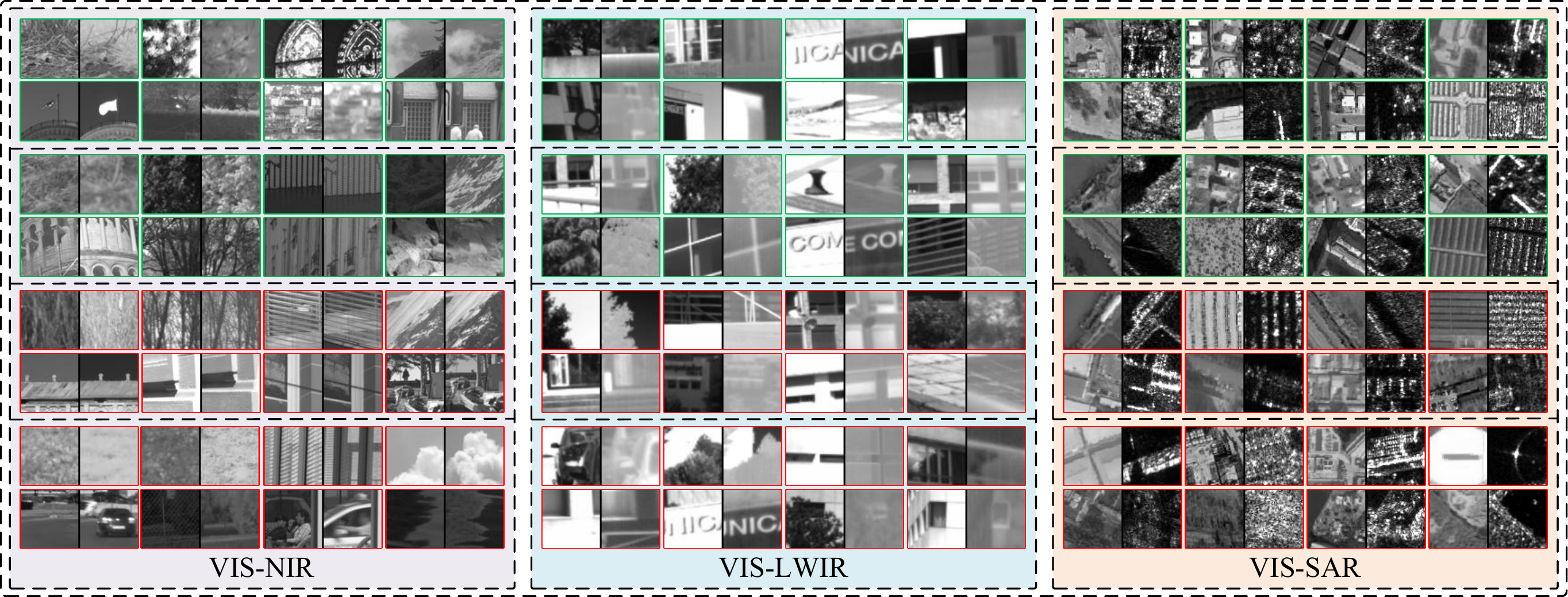}
    \vspace{-16pt}
    \caption{The matching effect of KGL-Net on the VIS-NIR patch dataset (VIS-NIR), VIS-LWIR patch dataset (VIS-LWIR) and OS patch dataset (VIS-SAR). Each cross-spectral scene is from top to bottom: correctly judged as matching, correctly judged as non-matching, misjudged as matching, and misjudged as non-matching. \textcolor{darkgreen}{Green} denotes a correct judgment, and \textcolor{darkred}{red} denotes an incorrect judgment.}
    \label{fig:fig09}
    \vspace{-4pt}
    
\end{figure*}

\subsection{Ablation Experiment}
\label{sec:ablation experiment}
\hlA{To fully validate the proposed KGL-Net, we perform detailed ablation experiments. More experiments can be found in the \hlB{\textit{Supplementary Material}}.}

\textbf{\textit{1) Break-down ablation.}} To further verify the proposed components, we conduct detailed ablation experiments in \Cref{tab:tab02}. Compared with the single metric network, the FPR95 decreases by 12.5\% (from 0.88 to 0.77), 35.2\% (from 0.88 to 0.57), and 38.6\% (from 0.88 to 0.54) when \textit{CNA}, \textit{\hlB{HNSM-M}}, and \textit{FGL} are added one by one. Among them, using our HNSM-M strategy to mine hard negative samples for learning results in the most significant improvement. Specifically, compared with using only the \textit{CNA}, the additional use of HNSM-M strategy resulted in a 26.0\% (from 0.77 to 0.57) decrease in FPR95. This is a significant improvement without increasing the parameters. In addition, from \cref{fig:fig07}, the curve of the final KGL-Net integrating all components is almost at the highest position and has the best AUC. At the same time, compared with \textit{``Our-w/ CNA''}, \textit{``Our-w/ CNA and HNSM-M''} with additional use of HNSM-M has a more significant AUC improvement. \hlA{Furthermore, the HNSM-M strategy still provides significant performance improvements even without using \textit{FGL}, indicating that \textit{FGL} is an enhancement strategy rather than a necessary condition for HNSM-M strategy to be effective. The primary role of the feature-guided loss is to provide a gentle alignment signal in the early stages of training, thereby stabilizing the joint optimization. It helps make the performance gains brought by the HNSM-M strategy more stable and substantial. \hlB{It is worth mentioning that the HNSM-M strategy is only performed during model training and introduces no learnable parameters. Both \textit{``KGL-Net''} and \textit{``KGL-Net w/o HNSM-M''} have 6.41 M parameters and a single-sample inference time of $0.23 \times 10^{-3}$ seconds. These results further verify the effectiveness and efficiency of the KGL-Net construction.}}

\begin{table*}[]
\caption{\hlA{Stability and significance analysis of FIL-Net and KGL-Net on multiple datasets. \textbf{CI} denotes the confidence interval. The \textit{p}-value is a significance metric, representing the probability that the difference between the two groups occurred by chance.}}
\vspace{-4pt}
\label{tab:tabs-stability-significance}
\setlength{\tabcolsep}{8pt}
\renewcommand{\arraystretch}{1.2}
\resizebox{\linewidth}{!}{
\begin{tabular}{c|ccc|ccc}
\hline
\multirow{2}{*}{Methods} & \multicolumn{3}{c|}{VIS-NIR patch dataset}                                                                                      & \multicolumn{3}{c}{VIS-LWIR patch dataset}                                                                                      \\ \cline{2-7} 
                         & \multicolumn{1}{c|}{Mean ± Variance ↓}        & \multicolumn{1}{c|}{95\% CI ↓}                     & \textit{p}-value  ↓         & \multicolumn{1}{c|}{Mean ± Variance ↓}        & \multicolumn{1}{c|}{95\% CI ↓}                     & \textit{p}-value  ↓        \\ \hline
FIL-Net                  & \multicolumn{1}{c|}{0.8600 ± 0.0495}          & \multicolumn{1}{c|}{{[}0.7985, 0.9215{]}}          & \multirow{2}{*}{$1.67 \times 10^{\scriptscriptstyle -6}$} & \multicolumn{1}{c|}{0.6960 ± 0.0378}          & \multicolumn{1}{c|}{{[}0.6490, 0.7430{]}}          & \multirow{2}{*}{$2.83 \times 10^{\scriptscriptstyle -5}$} \\ \cline{1-3} \cline{5-6}
KGL-Net                  & \multicolumn{1}{c|}{\textbf{0.5380 ± 0.0303}} & \multicolumn{1}{c|}{\textbf{{[}0.5003, 0.5757{]}}} &                            & \multicolumn{1}{c|}{\textbf{0.5080 ± 0.0319}} & \multicolumn{1}{c|}{\textbf{{[}0.4683, 0.5477{]}}} &                            \\ \hline
\multirow{2}{*}{Methods} & \multicolumn{3}{c|}{LLVIP patch dataset}                                                                                        & \multicolumn{3}{c}{OS patch dataset}                                                                                            \\ \cline{2-7} 
                         & \multicolumn{1}{c|}{Mean ± Variance ↓}        & \multicolumn{1}{c|}{95\% CI ↓}                     & \textit{p}-value ↓         & \multicolumn{1}{c|}{Mean ± Variance ↓}        & \multicolumn{1}{c|}{95\% CI ↓}                     & \textit{p}-value ↓         \\ \hline
FIL-Net                  & \multicolumn{1}{c|}{1.2500 ± 0.0620}          & \multicolumn{1}{c|}{{[}1.1730, 1.3270{]}}          & \multirow{2}{*}{$0.49 \times 10^{\scriptscriptstyle -6}$} & \multicolumn{1}{c|}{5.3660 ± 0.1050}          & \multicolumn{1}{c|}{{[}5.2356, 5.4964{]}}          & \multirow{2}{*}{$0.25 \times 10^{\scriptscriptstyle -6}$} \\ \cline{1-3} \cline{5-6}
KGL-Net                  & \multicolumn{1}{c|}{\textbf{0.7640 ± 0.0416}} & \multicolumn{1}{c|}{\textbf{{[}0.7124, 0.8156{]}}} &                            & \multicolumn{1}{c|}{\textbf{4.3160 ± 0.1041}} & \multicolumn{1}{c|}{\textbf{{[}4.1868, 4.4452{]}}} &                            \\ \hline
\end{tabular}
}
\vspace{-2pt}
\end{table*}

\begin{table}[]
\caption{\hlA{Performance of various methods under different generalization experimental settings. VIS-NIR denotes the VIS-NIR patch dataset, and VIS-LWIR denotes the VIS-LWIR patch dataset. The dataset on the left side of the arrow is used for training, and the one on the right side is used for testing.}}
\vspace{-4pt}
\label{tab:tabs-generalization}
\setlength{\tabcolsep}{4pt}
\renewcommand{\arraystretch}{1.2}
\resizebox{\linewidth}{!}{
\begin{tabular}{c|c|c}
\hline
Methods                 & VIS-LWIR → VIS-NIR    & VIS-NIR → VIS-LWIR    \\ \hline
SCFDM                   & 17.64 (-64.6\%)       & 30.90 (-77.6\%)       \\
AFD-Net                 & 18.26 (-65.8\%)       & 24.21 (-71.4\%)       \\
MR\_3A                  & 19.81 (-68.5\%)       & 28.06 (-75.3\%)       \\
MFD-Net                 & 17.25 (-63.8\%)       & 27.69 (-75.0\%)       \\
EFR-Net                 & {12.06 (-48.3\%)} & 25.63 (-73.0\%)       \\
FIL-Net                 & 19.83 (-68.5\%)       & {23.30 (-70.3\%)} \\ \hline
\rowcolor[HTML]{D7D7D7} 
\textbf{KGL-Net (Ours)} & \textbf{6.24}         & \textbf{6.92}         \\ \hline
\end{tabular}
}
\vspace{-4pt} 
\end{table}

\begin{figure}[!t]
\begin{minipage}{\columnwidth}
    \captionsetup{skip=4pt}
    \centering
    \includegraphics[width=\columnwidth]{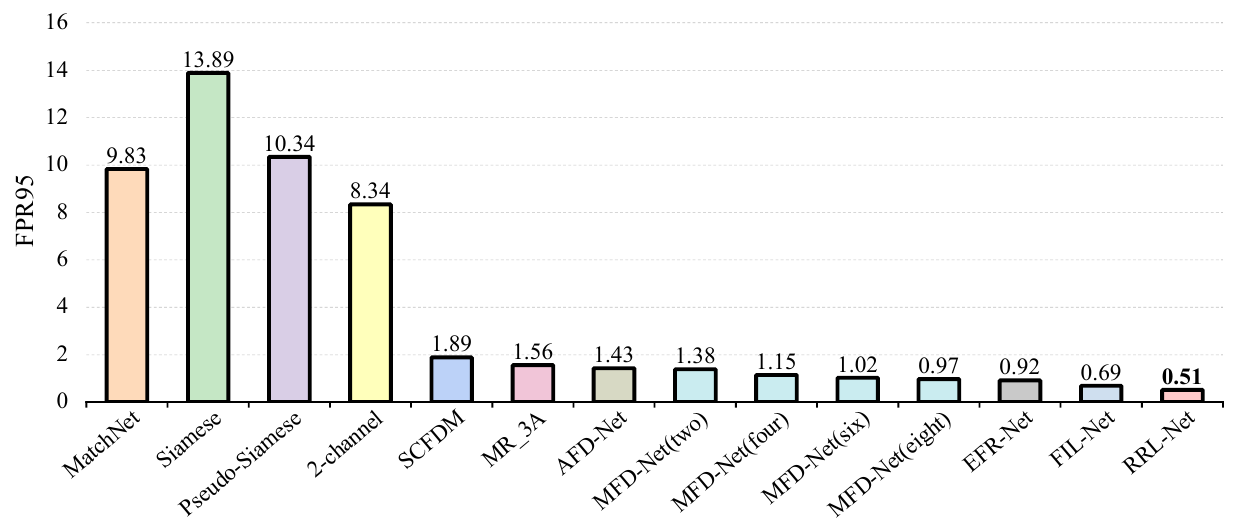}
     \vspace{-18pt}
    \caption{FPR95 comparison of multiple excellent methods on the VIS-LWIR patch dataset. \hlA{The number in parentheses is the branch count of MFD-Net.}}
    \label{fig:fig10}
\end{minipage}

 \vspace{13pt}

\begin{minipage}{\columnwidth}
    \captionsetup{skip=4pt}
    \centering
    \includegraphics[width=\columnwidth]{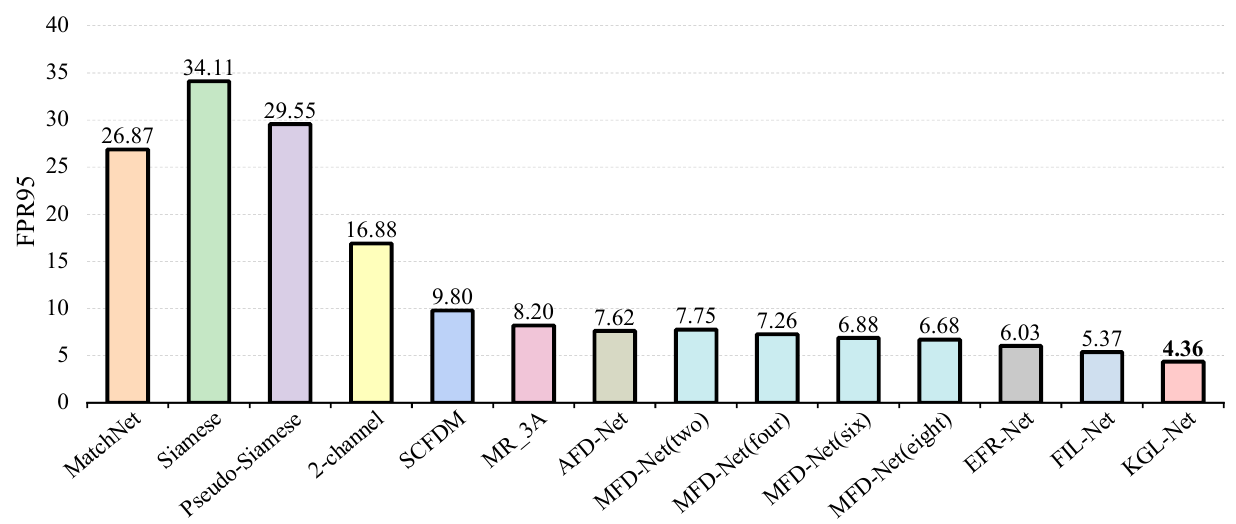}
    \vspace{-18pt}
    \caption{FPR95 comparison of multiple excellent methods on the OS patch dataset. \hlA{The number in parentheses is the branch count of MFD-Net.}}
    \label{fig:fig11}
    \vspace{-4pt} 
\end{minipage}

 \vspace{13pt}

\begin{minipage}{\columnwidth}
    \captionsetup{skip=4pt}
    \centering
    \includegraphics[width=0.8\columnwidth]{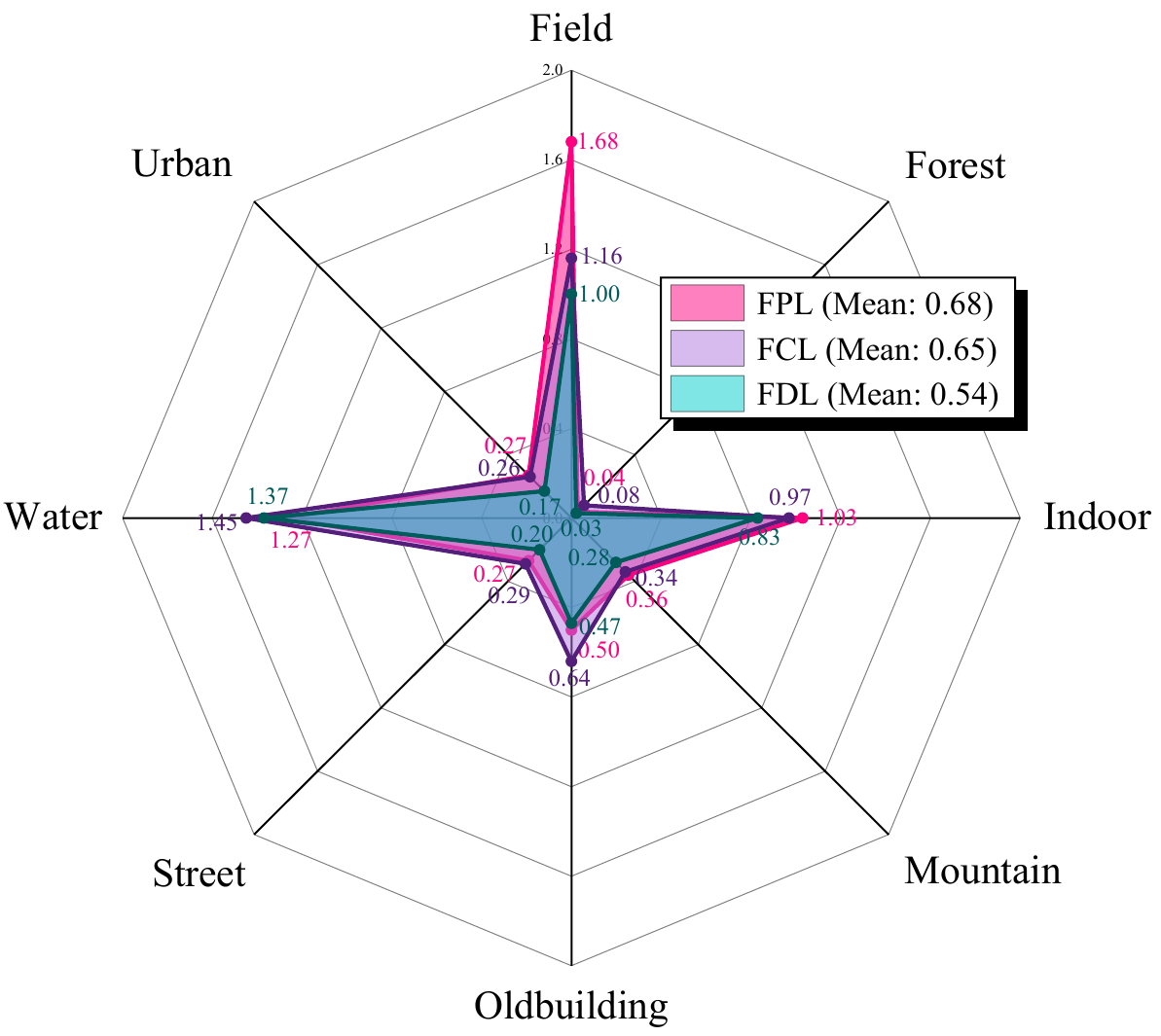}
    \vspace{-5pt}
    \caption{Performance comparison of different feature relationship learning methods in the metric branch on the VIS-NIR \hlB{patch} dataset. \textit{\textbf{FPL}} denotes feature product learning. \textit{\textbf{FCL}} denotes feature concatenation learning. \textit{\textbf{FDL}} denotes feature difference learning. \hlA{The numbers denote FPR95.}}
    \label{fig:fig12}
    \vspace{-8pt} 
\end{minipage}
\end{figure}

\textbf{\textit{2) Combined network architecture.}} To explore a stable and efficient bridge between descriptor learning and metric learning, we conduct a detailed experimental research on 20 combined network architectures. From \Cref{tab:tab03}, the construction of the combined network architecture has a significant impact. The \textit{A}, \textit{B} and \textit{D} series have poor overall performance, whereas the \textit{C} series have good overall performance. Owing to their own characteristics, compared with the metric network, the final output features of the feature extraction network part of the descriptor network focus more on global features. Directly sharing parameters between the two methods (\textit{B} series) or the parameters of the latter network layers (\textit{D} series) will cause the network to be overly constrained in the initial training stage, resulting in learning confusion and performance degradation. At the same time, compared with the two networks that do not share parameters at all (\textit{A} series), the sharing of the front part of the network \hlB{layers} (\textit{C} series) has better performance. This \hlB{shows} that the two methods sharing low-level network layers can help each other to extract richer detailed features, thereby promoting the performance improvement of the combined network architecture. It also further verifies that the descriptor network and the metric network need to establish appropriate connections. In addition, we find that when the descriptor network is a Siamese structure and the metric network is a Pseudo-Siamese structure (\textit{A3}, \textit{B3}, \textit{C3}, \textit{D3}), it has almost the best result in the same series. \hlB{These results provide useful guidance for the subsequent design of bridges.} Moreover, under the same combined network architecture, using the HNSM-M strategy reduces FPR95 by 6.5\%-39.5\%. This fully verifies the superiority of our HNSM-M strategy.

\begin{table*}[]
\caption{\hlB{Complexity and inference efficiency comparison of cross-spectral image patch matching methods from the past five years.}}
\vspace{-4pt}
\label{tab:tabs-complexity-efficiency}
\setlength{\tabcolsep}{3pt}
\renewcommand{\arraystretch}{1.1}
\resizebox{\linewidth}{!}{
\begin{tabular}{c|c|c|c|c|c|c|c|c}
\hline
Methods             & MFD-Net (two) & MFD-Net (four) & MFD-Net (six) & MFD-Net (eight) & SPIMNet & EFR-Net & FIL-Net &\textbf{ KGL-Net}      \\ \hline
Parameters (M)      & 34.71        & 81.31         & 116.27       & 151.24          & 130.89  & 9.17    & 33.07   & \textbf{6.41} \\ \hline
FLOPs (G)           & 2.09         & 4.20          & 6.29         & 8.38            & 10.58   & 1.79    & 4.09    & \textbf{1.76} \\ \hline
Inference time (ms) & 0.28         & 0.58          & 0.87         & 1.52            & 1.44    & 0.42    & 0.47    & \textbf{0.23} \\ \hline
\end{tabular}
}
\vspace{-4pt} 
\end{table*}

\begin{figure*}[]
    \captionsetup{skip=4pt}
    \centering
    \includegraphics[width=\textwidth]{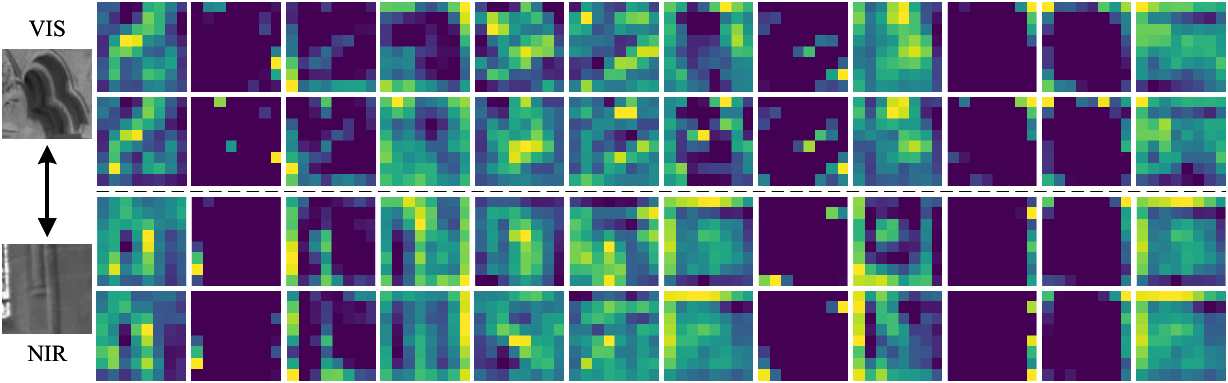}
    \vspace{-16pt} 
    \caption{The part channels of the highest layer feature map. The upper and lower rows show the outputs of the metric network, The middle two rows show the outputs of the descriptor network. Whether it is the upper two rows or the lower two rows, the corresponding feature maps have obvious similarities.}
    \label{fig:fig13}
    \vspace{-4pt}
\end{figure*}

To further explore the performance and robustness of the 20 combined network architectures, we also conducted experiments on the VIS-LWIR patch dataset. From \cref{fig:fig08}, it is basically consistent with the experimental conclusions on the VIS-NIR patch dataset. First, compared with other series, series \textit{C} has significantly better performance. At the same time, among the combined network architectures of the same series, the descriptor network is a Siamese structure and the metric network is a Pseudo-Siamese structure mostly have the best results. In addition, compared with not using the HNSM-M strategy, using the HNSM-M strategy has achieved significant performance improvement.

\textbf{\textit{3) Number of shared network layers.}} To more fully explore the performance differences between the descriptor network and the metric network under different sharing of low-level network layers, we explore various sharing \hlB{schemes} from the bottom to the top on the \textit{C3} architecture. \hlA{The experimental results are shown in \Cref{tab:tab04}. As the number of shared layers gradually increases, the FPR95 shows a decreasing trend followed by an increasing trend in both \hlB{the} VIS-NIR patch dataset and VIS-LWIR patch dataset. At the same time, the benefits of a reasonably shared low-level network layer are more evident for the VIS-LWIR scenario. On the VIS-NIR patch dataset, compared with not sharing and sharing 8 layers, sharing the first 4 layers will reduce FPR95 by 19.4\% (from 0.67 to 0.54) and 41.9\% (from 0.93 to 0.54). On the VIS-LWIR patch dataset, compared with not sharing and sharing 8 layers, sharing the first 4 layers will reduce FPR95 by 77.4\% (from 2.26 to 0.51) and 31.1\% (from 0.74 to 0.51).}

\hlA{\textit{\textbf{4) Feature-guided loss.}} We further explore the feature-guided loss in detail. First, we systematically investigate the two hyperparameters $\alpha$ and $\beta$. From \Cref{fig:figs-hyperparameters-fgl}, all FPR95 results are below 0.75 and there are no outliers. This demonstrates that the proposed KGL-Net is robust to both $\alpha$ and $\beta$. In particular, when $\alpha  \in [1,10]$ and $\beta  \in [0.1,1]$, the performance of KGL-Net is even more stable. At the same time, KGL-Net achieves optimal performance when $\alpha =1$ and $\beta =1$, with an FPR95 value of 0.54. Second, we experiment with different methods for calculating consistency in the feature-guided loss. In addition to the mean squared error (MSE) that we used, we also experiment with cosine similarity and KL divergence. According to experimental statistics, the final FPR95 values using cosine similarity and KL divergence are 0.55 and 0.57, respectively. Compared to using cosine similarity and KL divergence, using MSE results in a decrease in FPR95 by 1.8\% (from 0.55 to 0.54) and 5.3\% (from 0.57 to 0.54). In addition, we explore the use of projection heads with different settings before feature-guided loss. From \Cref{tab:tabs-projection-head}, compared to not using a projection head, using the projection heads does not bring about an improvement in performance, and may even lead to a slight decrease. Additionally, using the projection heads will result in extra training time.}

\subsection{Comparison with SOTA Methods}
\label{sec:comparison with sota methods}
We conduct a detailed comparison of our KGL-Net with various excellent image patch matching methods in three different cross-spectral scenarios (VIS-NIR, VIS-LWIR and VIS-SAR).

\begin{table}[!t]
\begin{minipage}{\columnwidth}
\caption{\hlA{Statistical analysis of the distributional similarity and directional consistency between the high-level features of the descriptor and metric networks in the KGL-Net.}}
\vspace{-5pt}
\label{tab:tabs-distributional-similarity}
\setlength{\tabcolsep}{3pt}
\renewcommand{\arraystretch}{1.2}
\resizebox{\columnwidth}{!}{
\begin{tabular}{c|cc|cc}
\hline
\multirow{2}{*}{Metric} & \multicolumn{2}{c|}{VIS Branch}              & \multicolumn{2}{c}{NIR Branch}              \\ \cline{2-5} 
                        & \multicolumn{1}{c|}{MMD ↓}   & CS ↑            & \multicolumn{1}{c|}{MMD ↓}   & CS ↑           \\ \hline
Field                   & \multicolumn{1}{c|}{0.1305} & 0.9900±0.0073 & \multicolumn{1}{c|}{0.1615} & 0.9896±0.0068 \\ \hline
Forest                  & \multicolumn{1}{c|}{0.1411} & 0.9916±0.0045 & \multicolumn{1}{c|}{0.1608} & 0.9917±0.0043 \\ \hline
Indoor                  & \multicolumn{1}{c|}{0.1999} & 0.9765±0.0120 & \multicolumn{1}{c|}{0.1943} & 0.9769±0.0138 \\ \hline
Mountain                & \multicolumn{1}{c|}{0.1416} & 0.9885±0.0069 & \multicolumn{1}{c|}{0.1505} & 0.9887±0.0068 \\ \hline
Oldbuilding             & \multicolumn{1}{c|}{0.1242} & 0.9855±0.0080 & \multicolumn{1}{c|}{0.1366} & 0.9858±0.0077 \\ \hline
Street                  & \multicolumn{1}{c|}{0.1015} & 0.9888±0.0078  & \multicolumn{1}{c|}{0.1110} & 0.9895±0.0071 \\ \hline
Urban                   & \multicolumn{1}{c|}{0.1364} & 0.9829±0.0101  & \multicolumn{1}{c|}{0.1416} & 0.9832±0.0096 \\ \hline
Water                   & \multicolumn{1}{c|}{0.1113} & 0.9883±0.0080  & \multicolumn{1}{c|}{0.1202} & 0.9894±0.0068 \\ \hline
\end{tabular}
}
\end{minipage}

\vspace{7pt}

\begin{minipage}{\columnwidth}
\caption{\hlA{Performance comparison of individual metric networks and combined networks for hard sample mining on the VIS-NIR patch dataset. PS-based metric: Pseudo-siamese-based metric network. S-based metric: Siamese-based metric network.}}
\vspace{-5pt}
\label{tab:tabs-individual-metric-networks}
\setlength{\tabcolsep}{9pt}
\renewcommand{\arraystretch}{1.2}
\resizebox{\columnwidth}{!}{
\begin{tabular}{c|c|c|c}
\hline
Scheme & PS-based metric & S-based metric & HNSM-M \\ \hline
FPR95  & 1.15 (-53.0\%)  & 1.00 (-46.0\%)  & 0.54   \\ \hline
\end{tabular}
}
\end{minipage}

\vspace{7pt}

\begin{minipage}{\columnwidth}
\caption{\hlA{Performance comparison of different sample mining strategies on the VIS-NIR patch dataset. SHM: Semi-Hard Mining; DWS: Distance-Weighted Sampling; MS: Multi-Similarity Mining.}}
\vspace{-5pt}
\label{tab:tabs-different-sample-mining}
\setlength{\tabcolsep}{5pt}
\renewcommand{\arraystretch}{1.2}
\resizebox{\columnwidth}{!}{
\begin{tabular}{c|c|c|c|c}
\hline
Strategy & SHM  & DWS  & MS   & HNSM-M \\ \hline
FPR95    & 0.57 (-5.3\%) & 0.61 (-11.5\%) & 0.59 (-8.5\%) & 0.54   \\ \hline
\end{tabular}
}
\end{minipage}
\vspace{1pt}
\end{table}

\begin{figure}[t]
  \captionsetup{skip=0pt}
  \centering
   \includegraphics[width=0.9\columnwidth]{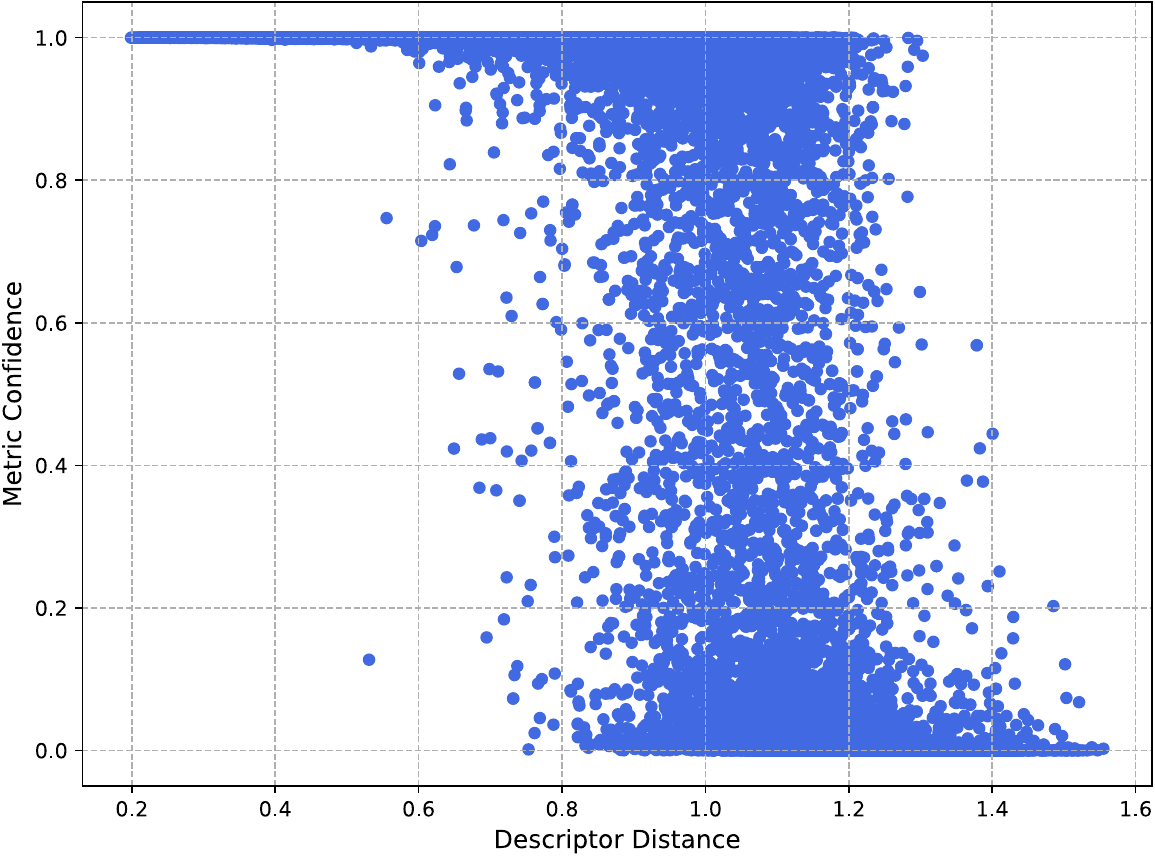}
   \caption{\hlA{Scatter distribution of Euclidean distances from the descriptor network versus confidence scores from the metric network on positive pairs in the ``Field'' test subset of the VIS-NIR patch dataset.}}
   \label{fig:figs-scatter}
   \vspace{-7pt}
\end{figure}

\textbf{\textit{1) VIS-NIR image matching.}} To verify the performance of KGL-Net in the VIS-NIR cross-spectral image patch matching scenario, we compare it with several excellent methods on the commonly used VIS-NIR patch dataset. From the results of ``Mean'' in \Cref{tab:tab01}, our KGL-Net achieves the best results. At the same time, it can be found that the performance improvement of cross-spectral image patch matching methods has gradually slowed down in recent years and performance bottlenecks have emerged. Compared with MFD-Net, the latest FIL-Net reduces the FPR95 by only 3.4\% (from 0.88 to 0.85). However, \hlB{compared with the latest FIL-Net, KGL-Net reduces FPR95 by 36.5\% (from 0.85 to 0.54) with fewer parameters. The improvement is enormous and breaks the existing inherent thinking that focuses on mining feature relations between image patches.} From the results of each subset, our KGL-Net achieves the best results on six subsets. Among them, the most significant improvement is achieved on the ``Field'' subset, which is identified as the most difficult subset of the dataset. This fully verifies that realizing hard negative sample mining for metric learning will effectively promote its correct discrimination of hard samples. In addition, from the results of ``VIS-NIR'' in \cref{fig:fig09}, most of the misjudged samples fall into two categories. On the one hand, owing to the spatiotemporal differences in the data acquisition, partially moving objects can cause image pairs to be marked as matching but visually non-matching. On the other hand, some non-matching image pairs with adjacent coordinates have extremely similar semantic features. In addition, from the correct sample pairs, our KGL-Net has an excellent matching effect in overcoming the spectral difference between VIS and NIR.

\textbf{\textit{2) VIS-LWIR image matching.}} To verify the performance of KGL-Net in the VIS-LWIR cross-spectral image patch matching scenario, we compare it with several excellent methods on the VIS-LWIR patch dataset. As presented in \cref{fig:fig10}, compared with other excellent methods, our KGL-Net can reduce the FPR95 by 26.1\% (from 0.69 to 0.51) - 96.3\% (from 13.89 to 0.51). \hlB{KGL-Net achieves a substantial improvement.} From the results of ``VIS-LWIR'' in \cref{fig:fig09}, compared with VIS-NIR, the pixel-level nonlinear difference in VIS-LWIR is more obvious. These results further verify that KGL-Net has excellent robustness in VIS-LWIR cross-spectral scenarios and datasets with a small number of samples.

\textbf{\textit{3) VIS-SAR image matching.}} To verify the performance of KGL-Net in the VIS-SAR cross-spectral image patch matching scenario, we compare it with several excellent methods on the OS patch dataset. From \cref{fig:fig11}, compared with other excellent methods, our KGL-Net can reduce FPR95 by 18.8\% (from 5.37 to 4.36) - 87.2\% (from 34.11 to 4.36), which verifies the excellent performance of our KGL-Net in VIS-SAR. From \cref{fig:fig09}, compared with VIS-NIR and VIS-LWIR, the VIS-SAR cross-spectral scenario has significantly more serious pixel-level nonlinear differences, which directly leads to the occurrence of misjudgments. However, it can be found from the correct sample pairs that KGL-Net also has good robustness in VIS-SAR cross-spectral scenarios.

\subsection{Analysis of Stability, Significance, and Generalization}
\label{sec:stability-significance-generalization}

\hlA{\textbf{\textit{1) Stability analysis.}} We perform five independent runs \hlB{on multiple datasets} and statistically analyze the mean, variance, and confidence intervals (CI). From \Cref{tab:tabs-stability-significance}, compared to FIL-Net, KGL-Net consistently shows significantly lower average FPR95 and variance across multiple datasets. In addition, based on the results of 95\% CI, KGL-Net is demonstrably more stable and superior. }

\hlA{\textbf{\textit{2) Significance analysis.}} We also conduct a significance analysis of FIL-Net and KGL-Net on multiple datasets. From the \textit{p}-values in the \Cref{tab:tabs-stability-significance}, the \textit{p}-values between the two methods are far less than 0.05 across all datasets, indicating that the improvement of KGL-Net compared to FIL-Net is significant and not due to chance. }

\hlA{\textbf{\textit{3) Generalization analysis.}} On the one hand, we train on the VIS-LWIR dataset and test on the VIS-NIR dataset. On the other hand, we train on the VIS-NIR dataset and test on the VIS-LWIR dataset. \hlB{From} \Cref{tab:tabs-generalization}, regardless of whether the setting is \textit{``VIS-LWIR → VIS-NIR''} or \textit{``VIS-NIR → VIS-LWIR''}, the proposed KGL-Net achieves the lowest FPR95. Specifically, compared to other methods in the \textit{``VIS-LWIR → VIS-NIR''} setting, the KGL-Net reduces the FPR95 by 48.3\% - 68.5\%. Compared to other methods in the \textit{``VIS-NIR → VIS-LWIR''} setting, the KGL-Net reduces the FPR95 by 70.3\% - 77.6\%. This fully validates that the proposed KGL-Net exhibits excellent generalization.}

\begin{table}[]
\caption{\hlA{Registration results of various excellent image patch matching methods on the VIS-NIR registration dataset.}}
\vspace{-4pt}
\label{tab:tabs-registration-results}
\setlength{\tabcolsep}{6pt}
\renewcommand{\arraystretch}{1.1}
\resizebox{\columnwidth}{!}{
\begin{tabular}{c|ccc|c|c}
\hline
                          & \multicolumn{3}{c|}{RMSE distribution}      &                                &                     \\
\multirow{-2}{*}{Methods} & \textless 1 ↑ & 1$\sim$5 & \textgreater 5 ↓ & \multirow{-2}{*}{$N_m$↑} & \multirow{-2}{*}{$I_{rr}$↑} \\ \hline
SCFDM                     & 121           & 29       & 10               & 150                            & 93.8\%              \\
AFD-Net                   & 121           & 29       & 10               & 150                            & 93.8\%              \\
MR\_3A                    & 118           & 31       & 11               & 149                            & 93.1\%              \\
MFD-Net                   & \underline{125}     & 30       & \underline{5}          & \underline{155}                      & \underline{96.9\%}        \\
EFR-Net                   & 124           & 30       & 6                & 154                            & 96.3\%              \\
FIL-Net                   & 115           & 38       & 7                & 153                            & 95.6\%              \\
RRL-Net                   & \textbf{127}  & 27       & 6                & 154                            & 96.3\%              \\ \hline
\rowcolor[HTML]{CFCECE} 
\textbf{KGL-Net (Ours)}                   & \underline{125}     & 31       & \textbf{4}       & \textbf{156}                   & \textbf{97.5\%}     \\ \hline
\end{tabular}
}
\end{table}

\begin{figure}[t]
  \captionsetup{skip=0pt}
  \centering
   \includegraphics[width=\columnwidth]{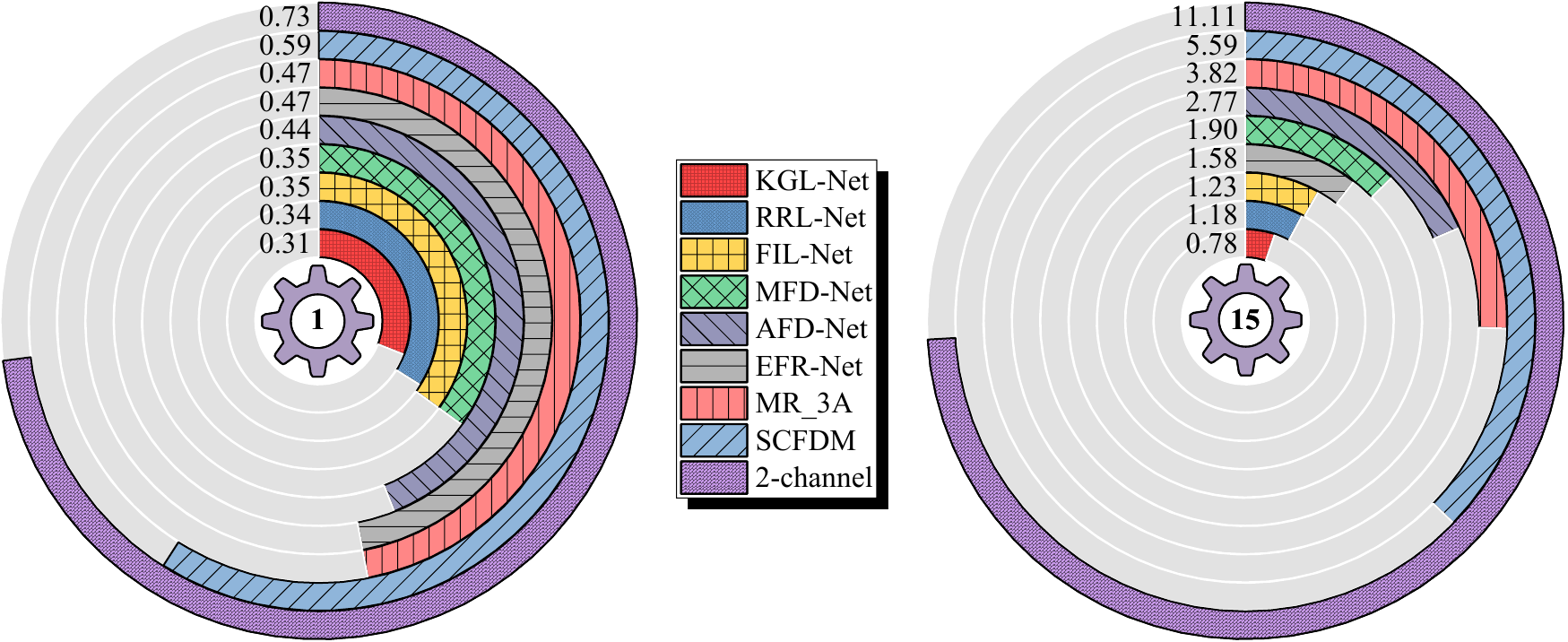}
   \caption{\hlA{FPR95 comparison of various excellent image patch matching methods on the two datasets. ``1'' and ``15'' denote the total value of one round. \textbf{Left:} the RIRE patch dataset. \textbf{Right:} the LLVIP patch dataset.}}
   \label{fig:figs-LLVIP-RIRE-results}
   \vspace{-4pt}
\end{figure}

\subsection{Discussion}
\label{sec:discussion}
\hlA{To fully validate the proposed KGL-Net, we conduct extensive discussions. More discussions can be found in the \hlB{\textit{Supplementary Material}}.}


\hlB{\textit{\textbf{1) Complexity and inference efficiency.}} From \Cref{tab:tabs-complexity-efficiency}, KGL-Net achieves the best performance in \textit{Parameters}, \textit{FLOPs}, and \textit{Inference time}. Specifically, compared with the latest FIL-Net, KGL-Net reduces \textit{Parameters} by 80.62\% (33.07 M to 6.41 M), \textit{FLOPs} by 56.97\% (4.09 G to 1.76 G), and \textit{Inference time} by 51.06\% (0.47 ms to 0.23 ms). Compared with the lightweight EFR-Net, KGL-Net reduces \textit{Parameters} by 30.10\% (9.17 M to 6.41 M), \textit{FLOPs} by 1.68\% (1.79 G to 1.76 G), and \textit{Inference time} by 45.2\% (0.42 ms to 0.23 ms). }

\textit{\textbf{2) Feature extraction consistency.}} As described in \Cref{sec:motivation}, we find that there is feature extraction consistency between metric learning based on feature difference learning and descriptor learning based on Euclidean distance. To further verify, we experiment with different feature relationship learning methods for the metric branch of the metric network. From \cref{fig:fig12}, compared with feature product learning focusing on consistent features and feature concatenation learning focusing on common features, feature difference learning can reduce Mean FPR95 by 20.6\% (from 0.68 to 0.54) and 16.9\% (from 0.65 to 0.54) respectively. At the same time, feature difference learning has the lowest FPR95 on most subsets, which verifies the rationality and robustness of feature difference learning in KGL-Net. In addition, we visualized the high-level feature maps output by the feature extraction network of the descriptor network and the metric network. From \cref{fig:fig13}, for the same image patch, the high-level features output by the descriptor branch and the metric branch are obviously consistent. \hlA{To further quantitatively assess the degree of alignment between the descriptor network and the metric network in the feature space within KGL-Net, we calculate two metrics: the maximum mean discrepancy (\textit{MMD}) and cosine similarity (\textit{CS}) for both the VIS and NIR branches. Specifically, \textit{MMD} measures the overall discrepancy between the feature distributions in the kernel space, where a smaller value indicates closer distributions. \textit{CS} evaluates the directional consistency of features corresponding to the same sample in the two networks, and a higher value reflects stronger semantic cooperation. From \Cref{tab:tabs-distributional-similarity}, the \textit{MMD} values across the subsets fall within the range of 0.10 to 0.20, with most values concentrated around 0.13, indicating that the two branches exhibit a generally aligned distribution in the kernel space. Moreover, the \textit{CS} for all subsets reaches 0.97 or above, implying that the descriptor and metric networks achieve an almost perfectly aligned feature direction in the semantic space. }


\begin{figure*}[]
    \captionsetup{skip=2pt}
    \centering
    \includegraphics[width=\textwidth]{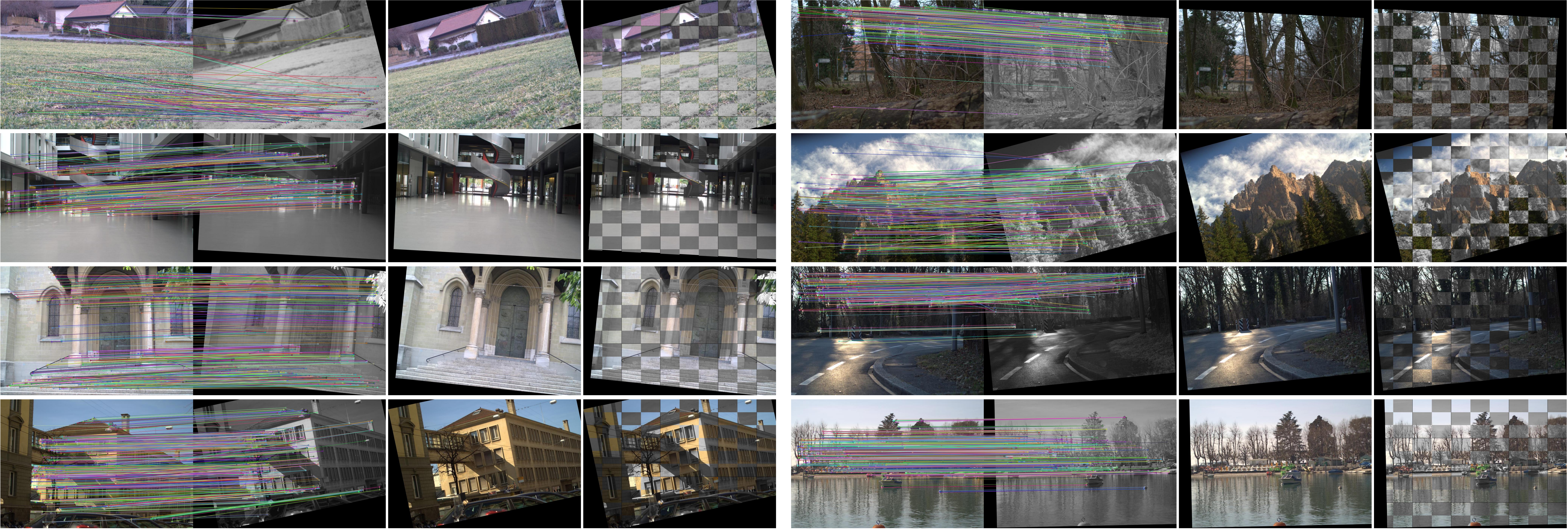}
    \vspace{-15pt}
    \caption{\hlA{Partial registration visualization of KGL-Net on the VIS-NIR registration dataset. Eight samples are from eight different subsets. For each sample, from left to right are the keypoint matching image, the registered VIS image, and the registered checkerboard image.}}
    \label{fig:figs-registration-visualization}
    \vspace{-4pt}
\end{figure*}

\begin{figure*}[]
    \captionsetup{skip=2pt}
    \centering
    \includegraphics[width=\textwidth]{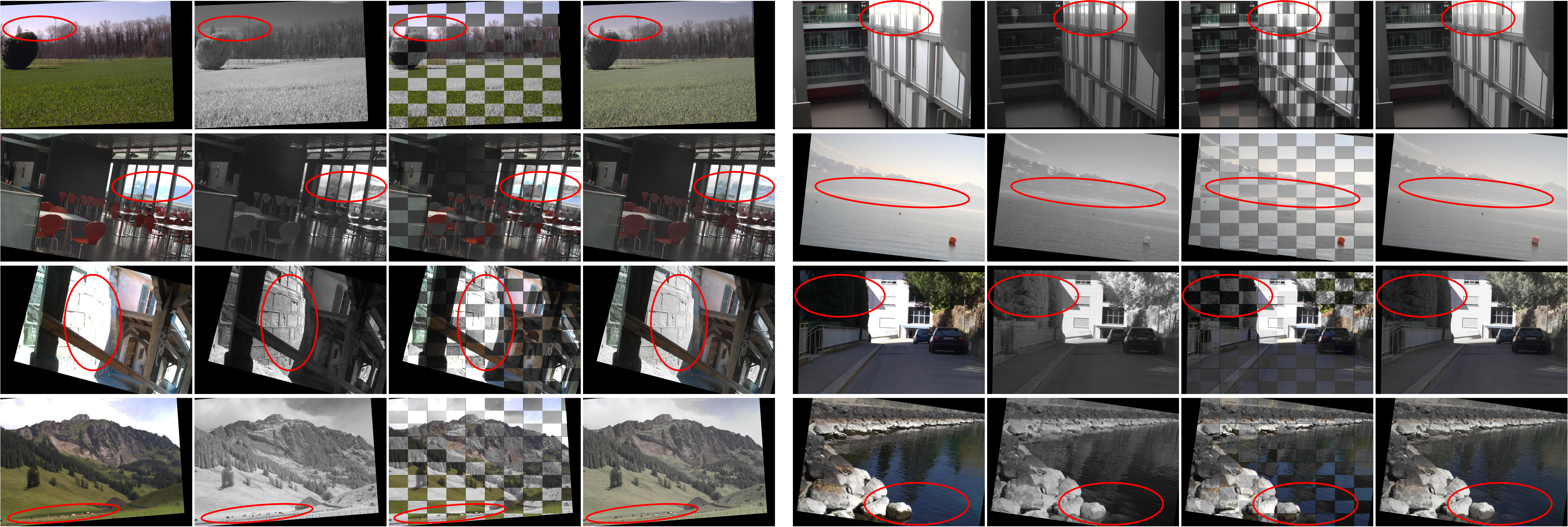}
    \vspace{-15pt}
    \caption{\hlA{Partial fusion visualization of KGL-Net on the VIS-NIR registration dataset. For each sample, from left to right are the VIS image, the NIR image, the registered checkerboard image, and the fused image.}}
    \label{fig:figs-fusion-visualization}
    \vspace{-5pt}
\end{figure*}

\textit{\textbf{\hlA{3) The rationality and effectiveness of the HNSM-M strategy.}}} First, on the positive pairs in the ``Field'' subset of the VIS-NIR patch dataset, we plot a scatter \hlB{diagram} as shown in \Cref{fig:figs-scatter}. A large number of points are concentrated in the upper left and lower right corners, and they show a negative correlation trend. Statistically, the Spearman rank correlation reaches 0.65, and the Pearson correlation reaches 0.59, both highly significant ($p < 1e^{-6}$). This indicates a stable monotonic relationship between descriptor distance and metric branch confidence. The greater the descriptor distance (higher difficulty), the lower the confidence output of the metric branch. Second, as shown in \Cref{tab:tab03}, when using random negative samples, the model performance will significantly decrease. This validates the effectiveness of using descriptor distance for selecting hard negative samples in the HNSM-M strategy. Third, we compare our method with the setting where hard negative mining is performed only on the metric network without descriptor network. When the individual metric network conducts hard negative mining, its architecture degenerates from C3 to B1 (Siamese-based) and B2 (Pseudo-siamese-based). From \Cref{tab:tabs-individual-metric-networks}, compared \hlB{with} the \textit{``PS-based metric''} and \textit{``S-based metric''}\hlB{, the HNSM-M strategy} reduce the FPR95 by 53.0\% and 46.0\%, respectively. This is mainly because the high-level features of metric learning networks focus on local similarity and are weaker in global semantic modeling, making it difficult to distinguish between semantically challenging samples and noisy negative samples. In contrast, HNSM-M incorporates structured semantic representations from the descriptor network, enabling the hard negative mining process to benefit jointly from the semantic distribution in the descriptor space and the confidence structure in the metric space. Finally, we experiment with replacing the hard sample selection in the descriptor branch of the HNSM-M strategy with several typical descriptor mining strategies (Semi-Hard Mining~\cite{schroff2015facenet}, Distance-Weighted Sampling~\cite{wu2017sampling}, Multi-Similarity Mining~\cite{wang2019multi}\hlB{)}. From \Cref{tab:tabs-different-sample-mining}, compared to using other strategies, using the HNSM-M strategy reduces the FPR95 by 5.3\% - 11.5\%, further validating the effectiveness of the HNSM-M strategy.

\textit{\textbf{4) Three ``bridges'' in KGL-Net.}} For the performance bottleneck of current cross-spectral image patch matching research, we break away from the inertial thinking of building a more complex feature relation extraction network structure and make the first attempt to explore a stable and efficient bridge between descriptor learning and metric learning. \hlB{To ensure the stability and efficiency of this combined network, KGL-Net fully considers the complementary advantages of descriptor learning and metric learning and comprehensively builds effective bridges at the front, middle, and back stages. Specifically, \textbf{firstly}, at the front stage, the low-level layers of the descriptor network and the metric network are shared to extract sufficient low-level detail features. \textbf{Secondly}, at the middle stage, the proposed feature-guided loss promotes the mutual guidance of features between the two branches and stabilizes joint optimization. \textbf{Finally}, at the back stage, the proposed HNSM-M strategy takes advantage of the descriptor to select the hard negative sample location index to guide metric learning to achieve hard negative sample mining.}

\subsection{Application}
\label{sec:application}

\hlA{\textbf{\textit{1) CT-MRI cross-modal image patch matching.}} We \hlB{compare KGL-Net with} several excellent methods on the RIRE patch dataset. The experimental results are shown in  \Cref{fig:figs-LLVIP-RIRE-results}. Compared to other methods, the proposed KGL-Net reduces the FPR95 by 8.8\% (from 0.34 to 0.31) - 57.5\% (from 0.73 to 0.31). Meanwhile, KGL-Net achieves an FPR95 of only 0.31. These results fully validate KGL-Net's excellent generalization performance in medical multimodal scenarios.}

\textbf{\textit{\hlA{2) Cross-spectral image patch matching under complex illumination conditions.}}} \hlA{We compare our KGL-Net with several excellent methods on the LLVIP patch dataset. The experimental results are shown in \Cref{fig:figs-LLVIP-RIRE-results}. Compared to other methods, the proposed KGL-Net reduces the FPR95 by 33.9\% (from 1.18 to 0.78) - 93.0\% (from 11.11 to 0.78). Meanwhile, KGL-Net achieves an FPR95 of only 0.78. In summary, KGL-Net achieves significant performance improvements in complex illumination conditions, further validating its excellent cross-modal matching capabilities.}

\hlA{\textbf{\textit{3) Cross-spectral image registration.}} We extend the KGL-Net to the cross-spectral image registration task and conduct experiments on the VIS-NIR registration dataset. Consistent with previous work, we first use the SIFT algorithm to extract 1000 key points from each image and extract 64×64 pixel local regions based on these key points as centers. Second, the image patches from different spectra are combined sequentially and then input into KGL-Net to obtain the similarity scores. Third, nearest neighbor matching and threshold filtering are used to identify candidate matching pairs. Finally, RANSAC is used to remove mismatches and estimate the matching parameters. From \Cref{tab:tabs-registration-results}, compared to other methods, KGL-Net has the largest $N_m$, indicating that KGL-Net has stronger robustness for complex scenarios. Additionally, \hlB{we} show some of the registration results in \Cref{fig:figs-registration-visualization}. From the registered checkerboard image, the VIS image after affine transformation almost completely overlaps with the NIR image, further validating the excellent performance of KGL-Net.}

\hlA{\textbf{\textit{4) Cross-spectral image fusion.}} Different spectral imaging techniques can capture scene information from complementary perceptual dimensions. Fusing multi-source images can obtain a more comprehensive and stable scene representation. Considering that misalignments often occur in real-world applications of cross-spectral images, we adopt a two-stage processing framework of first registration and then fusion. During the registration phase, we use a cross-spectral image registration method based on KGL-Net. During the fusion phase, a classic two-scale image fusion (TIF) strategy is employed. From \Cref{fig:figs-fusion-visualization}, VIS images are susceptible to extreme illumination conditions, leading to structural loss. While NIR images have stable contours, they lack semantic color information. The fusion result combines the complementary advantages of these two types of images, restoring the missing structures in the VIS image and adding color information to the NIR image, thereby significantly improving visual quality.}

\vspace{4pt}
\section{Conclusion}
\label{sec:conlusion}
\hlB{To address the performance bottleneck in current cross-spectral image patch matching}, we break away from the inertial thinking of building a more complex feature relation extraction network structure and make the first attempt to explore a stable and efficient bridge between descriptor learning and metric learning. On the basis of the discovered feature extraction consistency, we construct an efficient KGL-Net, which realizes the mining of hard negative samples for metric learning for the first time. Our KGL-Net achieves significant performance improvements while abandoning complex network structures. At the same time, we also conduct a detailed exploration of 20 basic combined network architectures, which provide \hlB{useful guidance} for the bridge design between the two methods. Extensive experiments demonstrate that the proposed KGL-Net consistently achieves state-of-the-art performance across various cross-spectral image patch matching scenarios (VIS-NIR, VIS-LWIR, VIS-SAR). We hope that our interesting findings and new research ideas can inspire researchers to rethink the feasibility and superiority of combined learning methods in cross-spectral image patch matching tasks.


\vspace{0pt}
\normalem
\bibliographystyle{IEEEtran} 
\bibliography{main}

@String(CVPR= {IEEE Conf. Comput. Vis. Pattern Recog.})

@String(ICCV= {Int. Conf. Comput. Vis.})

@String(ECCV= {Eur. Conf. Comput. Vis.})

@String(BMVC= {Brit. Mach. Vis. Conf.})

@String(ICIP = {IEEE Int. Conf. Image Process.})

@String(ACCV  = {ACCV})

@String(CVPRW= {IEEE Conf. Comput. Vis. Pattern Recog. Worksh.})

@String(CVPR  = {CVPR})

@String(ICCV  = {ICCV})

@String(ECCV  = {ECCV})

@String(BMVC  =	{BMVC})

@String(ICIP  = {ICIP})

@String(CVPRW= {CVPRW})

@article{yu2023feature,
  title={Feature interaction learning network for cross-spectral image patch matching},
  author={Yu, Chuang and Liu, Yunpeng and Zhao, Jinmiao and Wu, Shuhang and Hu, Zhuhua},
  journal={IEEE Transactions on Image Processing},
  volume={32},
  pages={5564--5579},
  year={2023},
  publisher={IEEE}
}

@article{quan2022deep,
  title={Deep feature correlation learning for multi-modal remote sensing image registration},
  author={Quan, Dou and Wang, Shuang and Gu, Yu and Lei, Ruiqi and Yang, Bowu and Wei, Shaowei and Hou, Biao and Jiao, Licheng},
  journal={IEEE Transactions on Geoscience and Remote Sensing},
  volume={60},
  pages={1--16},
  year={2022},
  publisher={IEEE}
}

@article{quan2022self,
  title={Self-distillation feature learning network for optical and SAR image registration},
  author={Quan, Dou and Wei, Huiyuan and Wang, Shuang and Lei, Ruiqi and Duan, Baorui and Li, Yi and Hou, Biao and Jiao, Licheng},
  journal={IEEE Transactions on Geoscience and Remote Sensing},
  volume={60},
  pages={1--18},
  year={2022},
  publisher={IEEE}
}

@inproceedings{lin2023zero,
  title={Zero-shot everything sketch-based image retrieval, and in explainable style},
  author={Lin, Fengyin and Li, Mingkang and Li, Da and Hospedales, Timothy and Song, Yi-Zhe and Qi, Yonggang},
  booktitle={Proceedings of the IEEE Conference on Computer Vision and Pattern Recognition (CVPR)},
  pages={23349--23358},
  year={2023}
}

@inproceedings{balntas2017hpatches,
  title={HPatches: A benchmark and evaluation of handcrafted and learned local descriptors},
  author={Balntas, Vassileios and Lenc, Karel and Vedaldi, Andrea and Mikolajczyk, Krystian},
  booktitle={Proceedings of the IEEE conference on computer vision and pattern recognition},
  pages={5173--5182},
  year={2017}
}

@inproceedings{shi2022beyond,
  title={Beyond cross-view image retrieval: Highly accurate vehicle localization using satellite image},
  author={Shi, Yujiao and Li, Hongdong},
  booktitle={Proceedings of the IEEE Conference on Computer Vision and Pattern Recognition (CVPR)},
  pages={17010--17020},
  year={2022}
}

@inproceedings{wang2018person,
  title={Person re-identification with cascaded pairwise convolutions},
  author={Wang, Yicheng and Chen, Zhenzhong and Wu, Feng and Wang, Gang},
  booktitle={Proceedings of the IEEE Conference on Computer Vision and Pattern Recognition (CVPR)},
  pages={1470--1478},
  year={2018}
}

@article{chen2022structure,
  title={Structure-aware positional transformer for visible-infrared person re-identification},
  author={Chen, Cuiqun and Ye, Mang and Qi, Meibin and Wu, Jingjing and Jiang, Jianguo and Lin, Chia-Wen},
  journal={IEEE Transactions on Image Processing},
  volume={31},
  pages={2352--2364},
  year={2022},
  publisher={IEEE}
}

@inproceedings{yang2019patch,
  title={Patch-based discriminative feature learning for unsupervised person re-identification},
  author={Yang, Qize and Yu, Hong-Xing and Wu, Ancong and Zheng, Wei-Shi},
  booktitle={Proceedings of the IEEE Conference on Computer Vision and Pattern Recognition (CVPR)},
  pages={3633--3642},
  year={2019}
}

@inproceedings{aguilera2016learning,
  title={Learning cross-spectral similarity measures with deep convolutional neural networks},
  author={Aguilera, Cristhian A and Aguilera, Francisco J and Sappa, Angel D and Aguilera, Cristhian and Toledo, Ricardo},
  booktitle={CVPRW},
  pages={1--9},
  year={2016}
}

@inproceedings{ofir2018deep,
  title={Deep multi-spectral registration using invariant descriptor learning},
  author={Ofir, Nati and Silberstein, Shai and Levi, Hila and Rozenbaum, Dani and Keller, Yosi and Bar, Sharon Duvdevani},
  booktitle={Proceedings of the IEEE International Conference on Image Processing (ICIP)},
  pages={1238--1242},
  year={2018},
  organization={IEEE}
}

@inproceedings{quan2019afd,
  title={AFD-Net: Aggregated feature difference learning for cross-spectral image patch matching},
  author={Quan, Dou and Liang, Xuefeng and Wang, Shuang and Wei, Shaowei and Li, Yanfeng and Huyan, Ning and Jiao, Licheng},
  booktitle={Proceedings of the IEEE International Conference on Computer Vision (ICCV)},
  pages={3017--3026},
  year={2019}
}

@article{quan2021element,
  title={Element-wise feature relation learning network for cross-spectral image patch matching},
  author={Quan, Dou and Wang, Shuang and Huyan, Ning and Chanussot, Jocelyn and Wang, Ruojing and Liang, Xuefeng and Hou, Biao and Jiao, Licheng},
  journal={IEEE Transactions on Neural Networks and Learning Systems},
  volume={33},
  number={8},
  pages={3372--3386},
  year={2021},
  publisher={IEEE}
}

@article{yu2022multibranch,
  title={Multibranch feature difference learning network for cross-spectral image patch matching},
  author={Yu, Chuang and Liu, Yunpeng and Li, Chenxi and Qi, Lin and Xia, Xin and Liu, Tianci and Hu, Zhuhua},
  journal={IEEE Transactions on Geoscience and Remote Sensing},
  volume={60},
  pages={1--15},
  year={2022},
  publisher={IEEE}
}

@article{yu2023efficient,
  title={Efficient feature relation learning network for cross-spectral image patch matching},
  author={Yu, Chuang and Zhao, Jinmiao and Liu, Yunpeng and Wu, Shuhang and Li, Chenxi},
  journal={IEEE Transactions on Geoscience and Remote Sensing},
  volume={61},
  pages={1--17},
  year={2023},
  publisher={IEEE}
}

@article{barnea1972class,
  title={A class of algorithms for fast digital image registration},
  author={Barnea, Daniel I and Silverman, Harvey F},
  journal={IEEE transactions on Computers},
  volume={100},
  number={2},
  pages={179--186},
  year={1972},
  publisher={IEEE}
}

@article{azriel1977coarse,
  title={Coarse-fine template matching},
  author={Azriel Rosenfeld and Gordon J. Vanderbrug},
  journal={IEEE Transactions on Systems, Man, and Cybernetics},
  volume={7},
  number={2},
  pages={104-107},
  year={1977},
  publisher={IEEE}
}

@article{vanderbrug1977two,
  title={Two-stage template matching},
  author={Vanderbrug and Rosenfeld},
  journal={IEEE transactions on computers},
  volume={100},
  number={4},
  pages={384--393},
  year={1977},
  publisher={IEEE}
}

@article{viola1997alignment,
  title={Alignment by maximization of mutual information},
  author={Viola, Paul and Wells III, William M},
  journal={International journal of computer vision},
  volume={24},
  number={2},
  pages={137--154},
  year={1997},
  publisher={Springer}
}

@article{lowe2004distinctive,
  title={Distinctive image features from scale-invariant keypoints},
  author={Lowe, David G},
  journal={International journal of computer vision},
  volume={60},
  pages={91--110},
  year={2004},
  publisher={Springer}
}

@inproceedings{dalal2005histograms,
  title={Histograms of oriented gradients for human detection},
  author={Dalal, Navneet and Triggs, Bill},
  booktitle={CVPR},
  volume={1},
  pages={886--893},
  year={2005},
  organization={Ieee}
}

@article{bay2008speeded,
  title={Speeded-up robust features (SURF)},
  author={Bay, Herbert and Ess, Andreas and Tuytelaars, Tinne and Van Gool, Luc},
  journal={Computer vision and image understanding},
  volume={110},
  number={3},
  pages={346--359},
  year={2008},
  publisher={Elsevier}
}

@inproceedings{rublee2011orb,
  title={ORB: An efficient alternative to SIFT or SURF},
  author={Rublee, Ethan and Rabaud, Vincent and Konolige, Kurt and Bradski, Gary},
  booktitle={Proceedings of the IEEE International Conference on Computer Vision (ICCV)},
  pages={2564--2571},
  year={2011},
  organization={Ieee}
}

@inproceedings{firmenichy2011multispectral,
  title={Multispectral interest points for RGB-NIR image registration},
  author={Firmenichy, Damien and Brown, Matthew and S{\"u}sstrunk, Sabine},
  booktitle={Proceedings of the IEEE International Conference on Image Processing (ICIP)},
  pages={181--184},
  year={2011},
  organization={IEEE}
}

@inproceedings{pinggera122012cross,
  title={On cross-spectral stereo matching using dense gradient features},
  author={Pinggera12, Peter and Breckon, Toby and Bischof, Horst},
  booktitle={Proceedings of the IEEE Conference on Computer Vision and Pattern Recognition (CVPR)},
  volume={2},
  pages={3},
  year={2012}
}

@article{aguilera2012multispectral,
  title={Multispectral image feature points},
  author={Aguilera, Cristhian and Barrera, Fernando and Lumbreras, Felipe and Sappa, Angel D and Toledo, Ricardo},
  journal={Sensors},
  volume={12},
  number={9},
  pages={12661--12672},
  year={2012},
  publisher={Molecular Diversity Preservation International (MDPI)}
}

@inproceedings{morris2007statistics,
  title={Statistics of infrared images},
  author={Morris, Nigel JW and Avidan, Shai and Matusik, Wojciech and Pfister, Hanspeter},
  booktitle={Proceedings of the IEEE Conference on Computer Vision and Pattern Recognition (CVPR)},
  pages={1--7},
  year={2007},
  organization={IEEE}
}

@inproceedings{alcantarilla2012kaze,
  title={KAZE features},
  author={Alcantarilla, Pablo Fern{\'a}ndez and Bartoli, Adrien and Davison, Andrew J},
  booktitle={Proceedings of the European Conference on Computer Vision (ECCV)},
  pages={214--227},
  year={2012},
  organization={Springer}
}

@article{mouats2014multispectral,
  title={Multispectral stereo odometry},
  author={Mouats, Tarek and Aouf, Nabil and Sappa, Angel Domingo and Aguilera, Cristhian and Toledo, Ricardo},
  journal={IEEE Transactions on Intelligent Transportation Systems},
  volume={16},
  number={3},
  pages={1210--1224},
  year={2014},
  publisher={IEEE}
}

@inproceedings{aguilera2015lghd,
  title={LGHD: A feature descriptor for matching across non-linear intensity variations},
  author={Aguilera, Cristhian A and Sappa, Angel D and Toledo, Ricardo},
  booktitle={Proceedings of the IEEE International Conference on Image Processing (ICIP)},
  pages={178--181},
  year={2015},
  organization={IEEE}
}

@article{lecun2015deep,
  title={Deep learning},
  author={LeCun, Yann and Bengio, Yoshua and Hinton, Geoffrey},
  journal={Nature},
  volume={521},
  number={7553},
  pages={436--444},
  year={2015},
  publisher={Nature Publishing Group UK London}
}

@inproceedings{simo2015discriminative,
  title={Discriminative learning of deep convolutional feature point descriptors},
  author={Simo-Serra, Edgar and Trulls, Eduard and Ferraz, Luis and Kokkinos, Iasonas and Fua, Pascal and Moreno-Noguer, Francesc},
  booktitle={Proceedings of the IEEE International Conference on Computer Vision (ICCV)},
  pages={118--126},
  year={2015}
}

@article{zhao2024hierarchical,
  title={Hierarchical equalization loss for long-tailed instance segmentation},
  author={Zhao, Yaochi and Chen, Sen and Liu, Shiguang and Hu, Zhuhua and Xia, Jingwen},
  journal={IEEE Transactions on Multimedia},
  volume={26},
  pages={6943--6955},
  year={2024},
  publisher={IEEE}
}

@article{chen2025discrete,
  title={Discrete Edge Feature Guided Rotation Detection Method for Remote Sensing Ship Wake},
  author={Chen, Junfei and Hu, Zhuhua and Zhao, Yaochi and Wu, Wei and Chen, Lihang and Hu, Yueming and Huang, Ba},
  journal={IEEE Journal of Selected Topics in Applied Earth Observations and Remote Sensing},
  year={2025},
  publisher={IEEE}
}

@article{hu2025cost,
  title={A Cost-Sensitive Small Vessel Detection Method for Maritime Remote Sensing Imagery},
  author={Hu, Zhuhua and Wu, Wei and Yang, Ziqi and Zhao, Yaochi and Xu, Lewei and Kong, Lingkai and Chen, Yunpei and Chen, Lihang and Liu, Gaosheng},
  journal={Remote Sensing},
  volume={17},
  number={14},
  pages={2471},
  year={2025},
  publisher={MDPI}
}

@article{zhao2025multi,
  title={Multi-scale direction-aware network for infrared small target detection},
  author={Zhao, Jinmiao and Shi, Zelin and Yu, Chuang and Liu, Yunpeng and Ying, Xinyi and Dai, Yimian},
  journal={IEEE Transactions on Geoscience and Remote Sensing},
  year={2025},
  publisher={IEEE}
}

@inproceedings{balntas2016learning,
  title={Learning local feature descriptors with triplets and shallow convolutional neural networks.},
  author={Balntas, Vassileios and Riba, Edgar and Ponsa, Daniel and Mikolajczyk, Krystian},
  booktitle={Proceedings of the British Machine Vision Conference (BMVC)},
  volume={1},
  number={2},
  pages={3},
  year={2016}
}

@article{balntas2016pn,
  title={PN-Net: Conjoined triple deep network for learning local image descriptors},
  author={Balntas, Vassileios and Johns, Edward and Tang, Lilian and Mikolajczyk, Krystian},
  journal={arXiv preprint arXiv:1601.05030},
  year={2016}
}

@inproceedings{kumar2016learning,
  title={Learning local image descriptors with deep siamese and triplet convolutional networks by minimising global loss functions},
  author={Kumar BG, Vijay and Carneiro, Gustavo and Reid, Ian},
  booktitle={Proceedings of the IEEE Conference on Computer Vision and Pattern Recognition (CVPR)},
  pages={5385--5394},
  year={2016}
}

@article{aguilera2017cross,
  title={Cross-spectral local descriptors via quadruplet network},
  author={Aguilera, Cristhian A and Sappa, Angel D and Aguilera, Cristhian and Toledo, Ricardo},
  journal={Sensors},
  volume={17},
  number={4},
  pages={873},
  year={2017},
  publisher={MDPI}
}

@inproceedings{zhang2017learning,
  title={Learning spread-out local feature descriptors},
  author={Zhang, Xu and Yu, Felix X and Kumar, Sanjiv and Chang, Shih-Fu},
  booktitle={Proceedings of the IEEE International Conference on Computer Vision (ICCV)},
  pages={4595--4603},
  year={2017}
}

@article{mishchuk2017working,
  title={Working hard to know your neighbor's margins: Local descriptor learning loss},
  author={Mishchuk, Anastasiia and Mishkin, Dmytro and Radenovic, Filip and Matas, Jiri},
  journal={NeurIPS},
  volume={30},
  year={2017}
}

@inproceedings{wang2017train,
  title={How to train triplet networks with 100k identities?},
  author={Wang, Chong and Zhang, Xue and Lan, Xipeng},
  booktitle={ICCVW},
  pages={1907--1915},
  year={2017}
}

@inproceedings{zheng2019hardness,
  title={Hardness-aware deep metric learning},
  author={Zheng, Wenzhao and Chen, Zhaodong and Lu, Jiwen and Zhou, Jie},
  booktitle={Proceedings of the IEEE Conference on Computer Vision and Pattern Recognition (CVPR)},
  pages={72--81},
  year={2019}
}

@inproceedings{wang2019better,
  title={Better and faster: Exponential loss for image patch matching},
  author={Wang, Shuang and Li, Yanfeng and Liang, Xuefeng and Quan, Dou and Yang, Bowu and Wei, Shaowei and Jiao, Licheng},
  booktitle={Proceedings of the IEEE International Conference on Computer Vision (ICCV)},
  pages={4812--4821},
  year={2019}
}

@inproceedings{tian2017l2,
  title={L2-net: Deep learning of discriminative patch descriptor in euclidean space},
  author={Tian, Yurun and Fan, Bin and Wu, Fuchao},
  booktitle={Proceedings of the IEEE Conference on Computer Vision and Pattern Recognition (CVPR)},
  pages={661--669},
  year={2017}
}

@inproceedings{tian2019sosnet,
  title={Sosnet: Second order similarity regularization for local descriptor learning},
  author={Tian, Yurun and Yu, Xin and Fan, Bin and Wu, Fuchao and Heijnen, Huub and Balntas, Vassileios},
  booktitle={Proceedings of the IEEE Conference on Computer Vision and Pattern Recognition (CVPR)},
  pages={11016--11025},
  year={2019}
}

@article{tian2020hynet,
  title={HyNet: Learning local descriptor with hybrid similarity measure and triplet loss},
  author={Tian, Yurun and Barroso Laguna, Axel and Ng, Tony and Balntas, Vassileios and Mikolajczyk, Krystian},
  journal={NeurIPS},
  volume={33},
  pages={7401--7412},
  year={2020}
}

@inproceedings{han2015matchnet,
  title={Matchnet: Unifying feature and metric learning for patch-based matching},
  author={Han, Xufeng and Leung, Thomas and Jia, Yangqing and Sukthankar, Rahul and Berg, Alexander C},
  booktitle={Proceedings of the IEEE Conference on Computer Vision and Pattern Recognition (CVPR)},
  pages={3279--3286},
  year={2015}
}

@inproceedings{zagoruyko2015learning,
  title={Learning to compare image patches via convolutional neural networks},
  author={Zagoruyko, Sergey and Komodakis, Nikos},
  booktitle={Proceedings of the IEEE Conference on Computer Vision and Pattern Recognition (CVPR)},
  pages={4353--4361},
  year={2015}
}

@article{quan2021multi,
  title={Multi-relation attention network for image patch matching},
  author={Quan, Dou and Wang, Shuang and Li, Yi and Yang, Bowu and Huyan, Ning and Chanussot, Jocelyn and Hou, Biao and Jiao, Licheng},
  journal={IEEE Transactions on Image Processing},
  volume={30},
  pages={7127--7142},
  year={2021},
  publisher={IEEE}
}

@inproceedings{quan2018cross,
  title={Cross-spectral image patch matching by learning features of the spatially connected patches in a shared space},
  author={Quan, Dou and Fang, Shuai and Liang, Xuefeng and Wang, Shuang and Jiao, Licheng},
  booktitle={Proceedings of the Asian Conference on Computer Vision (ACCV)},
  pages={115--130},
  year={2018},
  organization={Springer}
}

@inproceedings{en2018ts,
  title={TS-NET: Combining modality specific and common features for multimodal patch matching},
  author={En, Sovann and Lechervy, Alexis and Jurie, Fr{\'e}d{\'e}ric},
  booktitle={Proceedings of the IEEE International Conference on Image Processing (ICIP)},
  pages={3024--3028},
  year={2018},
  organization={IEEE}
}

@article{ko2023spectral,
  title={Spectral-invariant matching network},
  author={Ko, Yeongmin and Jang, Yong-Jun and Dinh, Vinh Quang and Jeon, Hae-Gon and Jeon, Moongu},
  journal={Information Fusion},
  volume={91},
  pages={623--632},
  year={2023},
  publisher={Elsevier}
}

@article{kingma2014adam,
  title={Adam: A method for stochastic optimization},
  author={Kingma, Diederik P and Ba, Jimmy},
  journal={arXiv preprint arXiv:1412.6980},
  year={2014}
}

@inproceedings{singh2020filter,
  title={Filter response normalization layer: Eliminating batch dependence in the training of deep neural networks},
  author={Singh, Saurabh and Krishnan, Shankar},
  booktitle={Proceedings of the IEEE Conference on Computer Vision and Pattern Recognition (CVPR)},
  pages={11237--11246},
  year={2020}
}

@inproceedings{wang2020eca,
  title={ECA-Net: Efficient channel attention for deep convolutional neural networks},
  author={Wang, Qilong and Wu, Banggu and Zhu, Pengfei and Li, Peihua and Zuo, Wangmeng and Hu, Qinghua},
  booktitle={Proceedings of the IEEE Conference on Computer Vision and Pattern Recognition (CVPR)},
  pages={11534--11542},
  year={2020}
}

@article{yu2025relational,
  title={Relational Representation Learning Network for Cross-Spectral Image Patch Matching},
  author={Yu, Chuang and Liu, Yunpeng and Zhao, Jinmiao and Quan, Dou and Shi, Zelin and Yue, Xiangyu},
  journal={Information Fusion},
  pages={103749},
  year={2025},
  publisher={Elsevier}
}

@inproceedings{schroff2015facenet,
  title={Facenet: A unified embedding for face recognition and clustering},
  author={Schroff, Florian and Kalenichenko, Dmitry and Philbin, James},
  booktitle={Proceedings of the IEEE conference on computer vision and pattern recognition},
  pages={815--823},
  year={2015}
}

@inproceedings{wu2017sampling,
  title={Sampling matters in deep embedding learning},
  author={Wu, Chao-Yuan and Manmatha, R and Smola, Alexander J and Krahenbuhl, Philipp},
  booktitle={Proceedings of the IEEE international conference on computer vision},
  pages={2840--2848},
  year={2017}
}

@inproceedings{wang2019multi,
  title={Multi-similarity loss with general pair weighting for deep metric learning},
  author={Wang, Xun and Han, Xintong and Huang, Weilin and Dong, Dengke and Scott, Matthew R},
  booktitle={Proceedings of the IEEE/CVF conference on computer vision and pattern recognition},
  pages={5022--5030},
  year={2019}
}

@article{passov2026multi,
  title={Multi-Sensor Matching with HyperNetworks},
  author={Passov, Eli and Netanyahu, Nathan S and Keller, Yosi},
  journal={arXiv preprint arXiv:2601.12325},
  year={2026}
}

@article{zhang2025mssi,
  title={MSSI: a multi-scale shift-invariant metric learning network for visible and long-wave infrared image patch matching},
  author={Zhang, Xiuwei and Li, Yanping and Qi, Zhaoshuai and Liang, Yue and Sun, Yi},
  journal={International Journal of Remote Sensing},
  volume={46},
  number={19},
  pages={7077--7103},
  year={2025},
  publisher={Taylor \& Francis}
}

\vfill

\end{document}